%% file: egpaper_for_review.tex
\documentclass[10pt,twocolumn,letterpaper]{article}

\usepackage{cvpr}
\usepackage{times}
\usepackage{epsfig}
\usepackage{graphicx}
\usepackage{amsmath}
\usepackage{amssymb}
\usepackage{enumerate}
\usepackage[shortlabels]{enumitem}    
\usepackage{float}

\usepackage{times}
\usepackage{multirow}
\usepackage{epsfig}
\usepackage{graphicx}
\usepackage{amsmath}
\usepackage{amssymb}
\usepackage{color,soul}
\usepackage{sidecap}
\usepackage{booktabs}
\usepackage{adjustbox}
\usepackage{float}

\input{def}

\usepackage{multirow}
\usepackage{subcaption} 
\usepackage{caption}
\usepackage{multicol}
\usepackage{algorithm} 
\usepackage{algpseudocode} 

\usepackage[title]{appendix}

\usepackage[numbers,sort]{natbib} 

\usepackage[pagebackref=true,breaklinks=true,letterpaper=true,colorlinks,bookmarks=false]{hyperref}

\DeclareMathAlphabet{\mathpzc}{OT1}{pzc}{m}{it}

\providecommand{\eg}[0]{e.g\xperiod}
\providecommand{\ie}[1]{i.e\xperiod}
\providecommand{\etal}[2]{et al.\xperiod}
\cvprfinalcopy 

\def\cvprPaperID{611} 

\ifcvprfinal\fi
\begin{document}

\title{MAST: A Memory-Augmented Self-Supervised Tracker}

\author{Zihang Lai \quad\quad\quad\quad Erika Lu \quad\quad\quad\quad Weidi Xie\\
Visual Geometry Group, Department of Engineering Science\\
University of Oxford\\
{\tt\small \{zlai, erika, weidi\}@robots.ox.ac.uk}
}

\maketitle

\begin{abstract}
Recent interest in self-supervised dense tracking has yielded rapid progress, 
but performance still remains far from supervised methods.
We propose a dense tracking model trained on videos \textbf{without any annotations} that surpasses
previous self-supervised methods on existing benchmarks by a significant margin ($+15\%$), 
and achieves performance comparable to supervised methods.
In this paper, we first reassess the traditional choices used for self-supervised training and reconstruction loss 
by conducting thorough experiments that finally elucidate the optimal choices. 
Second, we further improve on existing methods by augmenting our architecture with a crucial memory component. 
Third, we benchmark on large-scale semi-supervised video object segmentation~(\emph{aka.}~dense tracking), 
and propose a new metric: generalizability.
Our first two contributions yield a self-supervised network that for the \textbf{first time} is competitive 
with supervised methods on standard evaluation metrics of dense tracking.
When measuring generalizability, 
we show self-supervised approaches are actually \emph{superior} to the majority of supervised methods.
We believe this new generalizability metric can better capture the real-world use-cases for dense tracking, 
and will spur new interest in this research direction.
Our code will be released at \url{https://github.com/zlai0/MAST}.
\end{abstract}

\section{Introduction}
\input{sec/intro_v3.tex}
\input{sec/related.tex}
\input{sec/method.tex}
\input{sec/imp.tex}

\input{sec/exp.tex}
\section{Conclusion}
In summary, 
we present a memory-augmented self-supervised model that enables accurate and generalizable pixel-level tracking.
The algorithm is trained without any semantic annotation, 
and surpasses previous self-supervised methods on existing benchmarks by a significant margin,
narrowing the gap with supervised methods. 
On unseen object categories, 
our model actually outperforms all but one existing methods that are trained with heavy supervision.
As computation power grows and more high quality videos become available, 
we believe that self-supervised learning algorithms 
can serve as a strong competitor to their supervised counterparts for their flexibility and generalizability.

\section{Acknowledgements}
\noindent
The authors would like to thank Andrew Zisserman for helpful discussions,
Olivia Wiles, Shangzhe Wu, Sophia Koepke and Tengda Han for proofreading.
Financial support for this project is provided by EPSRC Seebibyte Grant EP/M013774/1.
Erika Lu is funded by the Oxford-Google DeepMind Graduate Scholarship.

{\small
\bibliographystyle{ieee_fullname}
\bibliography{shortstrings,vgg_local,vgg_other,egbib}
}

\input{supp_sec.tex}
\end{document}

%% file: def.tex
\newcommand{\comment}[1]{}

\usepackage{pifont}
\newcommand{\cmark}{\ding{51}}%
\newcommand{\xmark}{\ding{55}}%

%% file: sec/intro_v3.tex
\begin{figure}[t]
\includegraphics[width=0.45\textwidth]{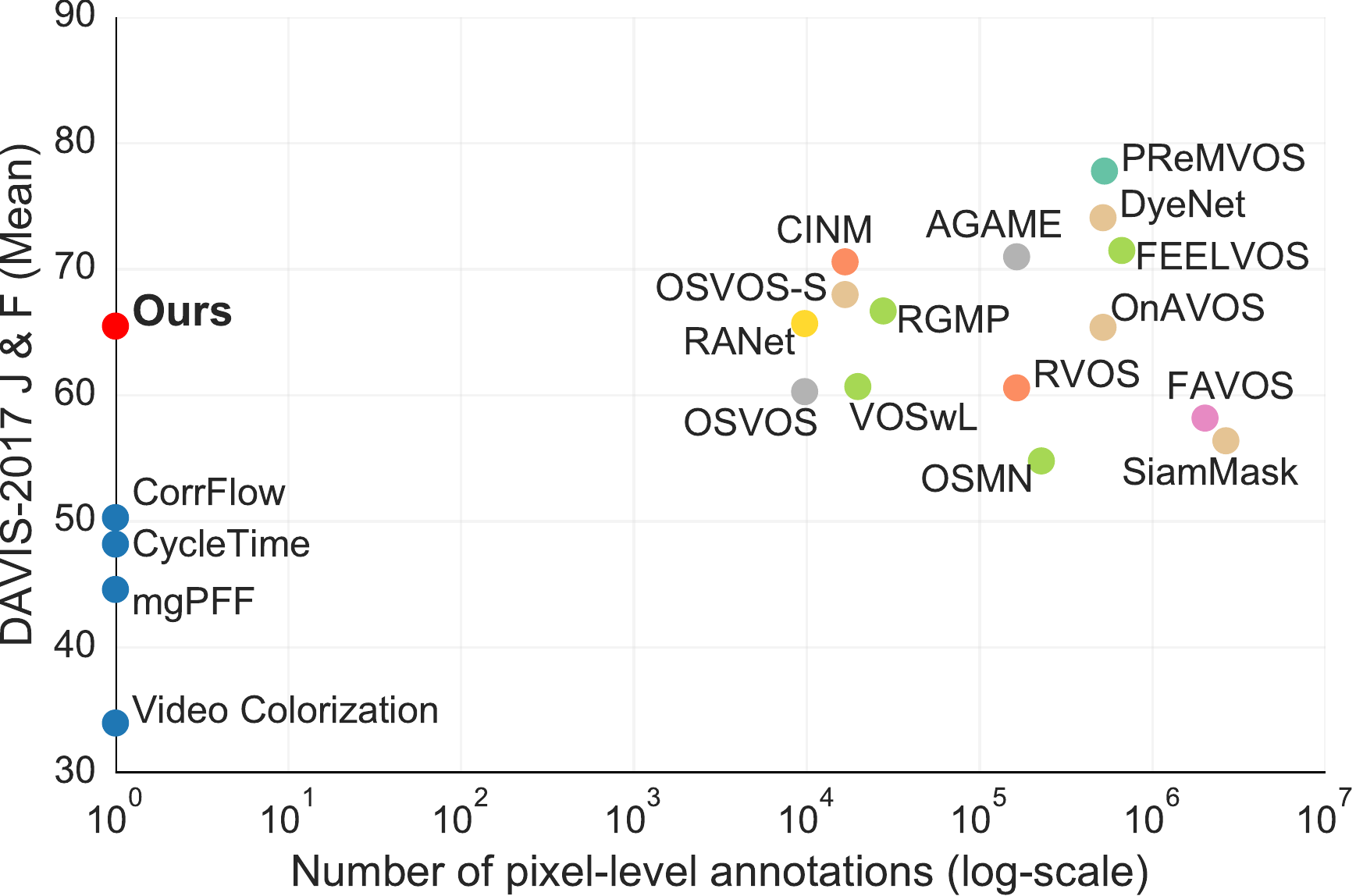}
\centering
\vspace{-0.25cm}
\caption{Comparison with other recent works on the DAVIS-2017 benchmarks,
\ie dense tracking or semi-supervised video segmentation given the first frame annotation.
The proposed approach significantly outperforms the other self-supervised approaches, 
and even comparable to approaches trained with heavy supervision on ImageNet, COCO, Pascal, DAVIS, Youtube-VOS. 
In the x-axis, we only count pixel-wise segmentation.\\
\textbf{Notation}:
CINM~\cite{Bao18},
OSVOS~\cite{Caelles17}, 
AGAME~\cite{Johnander19},
VOSwL~\cite{Khoreva18},
FAVOS~\cite{Cheng18},
mgPFF~\cite{Song19}, 
CorrFlow~\cite{Lai19}, 
DyeNet~\cite{Li18ECCV}, 
PReMVOS~\cite{Luiten18}.
OSVOS-S~\cite{Maninis18}, 
RGMP~\cite{Oh18},
RVOS~\cite{Ventura19},
FEELVOS~\cite{Voigtlaender19}, 
OnAVOS~\cite{Voigtlaender17}, 
Video Colorization~\cite{Vondrick18},
SiamMask~\cite{Wang19a},
CycleTime~\cite{Wang19}, 
RANet~\cite{Wang19RA},
OSMN~\cite{Yang18}, 
}
\label{fig:teaser}
\vspace{-15pt}
\end{figure}

\begin{figure*}[!htb]
  \centering
  \includegraphics[width=.95\textwidth]{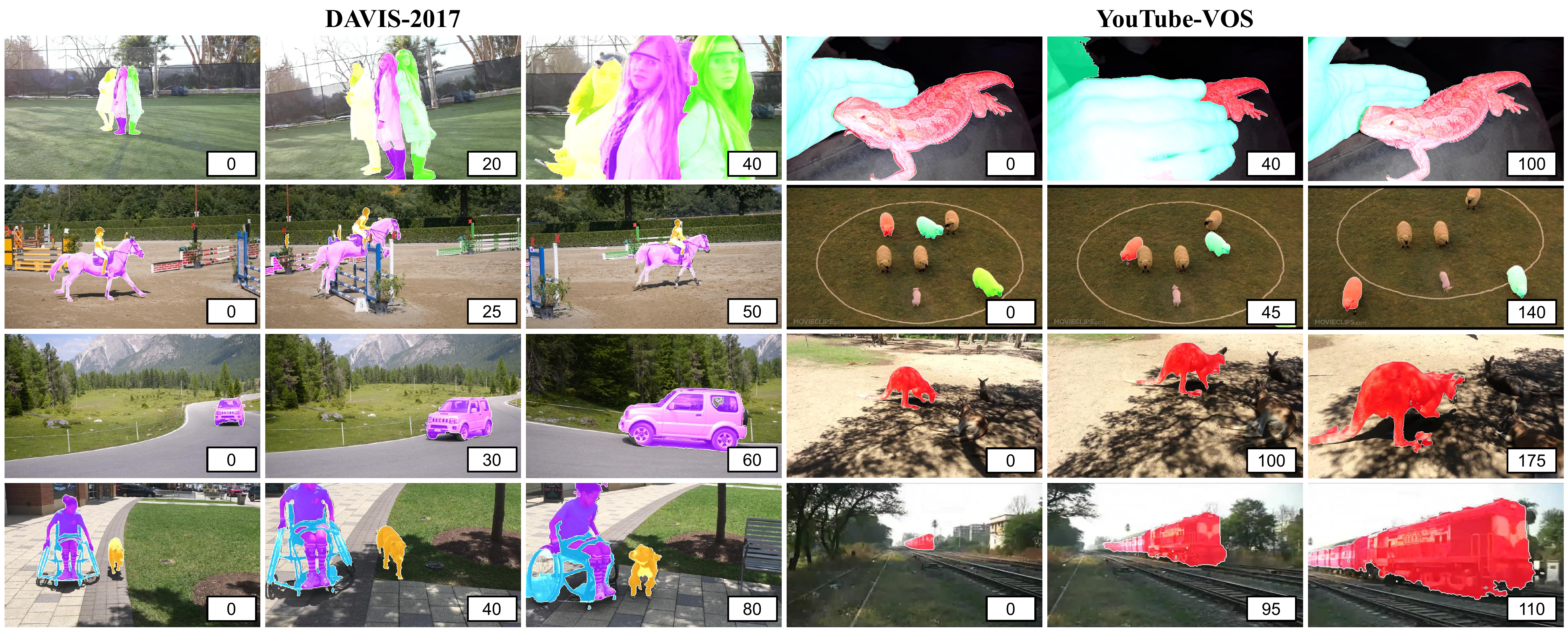}
  \vspace{-3pt}
  \caption{\textbf{Train once, test on multiple datasets:}
Qualitative results from our \emph{self-supervised dense tracking model} on DAVIS-2017 and YouTube-VOS dataset. 
The number on the top left refers to the frame number in the video.
For all examples, the mask of the $0th$ frame is given, and the task is to track the objects along with the video.
Our self-supervised tracking model is able to deal with challenging scenarios, 
such as large camera motion, occlusion and disocclusion, large deformation and scale variation.}
\label{fig:qualitative}
\vspace{-15pt}
\end{figure*}

Although the working mechanisms of the human visual system remain somewhat obscure at the level of neurophysiology, 
it is a consensus that tracking objects 
is a fundamental ability that a baby starts developing at two to three months of age~\cite{Brazelton66, Kremenitzer79, Hofsten82}. 
Similarly, in computer vision systems, 
tracking plays key roles in many applications ranging from autonomous driving to video surveillance. 

Given arbitrary objects defined in the first frame, 
a tracking algorithm aims to relocate the same object throughout the entire video sequence. 
In the literature, 
tracking can be cast into two categories:
the first is Visual Object Tracking (VOT)~\cite{VOT_TPAMI}, 
where the goal is to relocalize objects with bounding boxes throughout the video;
the other aims for more fine-grained tracking, 
\ie~relocalize the objects with pixel-level segmentation masks,
also known as Semi-supervised Video Object Segmentation (Semi-VOS)~\cite{Pont-Tuset17}.
In this paper, 
we focus on the latter case, 
and will refer to it interchangeably with \emph{dense tracking} from here on. 

In order to train such dense tracking systems, 
most recent approaches rely on supervised training with extensive human annotations~(see Figure~\ref{fig:teaser}).
For instance, an ImageNet~\cite{Deng09} pre-trained ResNet~\cite{He16} is typically adopted as a feature encoder,
and further fine-tuned on images or video frames annotated with fine-grained, pixelwise segmentation masks, 
\eg COCO~\cite{Lin14}, Pascal~\cite{Everingham09a}, DAVIS~\cite{Pont-Tuset17} and YouTube-VOS~\cite{Xu18}.
Despite their success, 
this top-down training scheme seems counter-intuitive when considering the development of the human visual system,
as infants can track and follow slow-moving objects before they are able to map objects to semantic meanings. 
With this evidence, 
it is unlikely the case that humans develop their tracking ability in a top-down manner~(supervised by semantics), 
at least not at the early-stage development of the visual system.

In contrast to the aforementioned approaches based on heavy supervision, 
self-supervised methods~\cite{Vondrick18, Wang19, Lai19, Wang19_Tracking} have recently been introduced, 
leading to more neurophysiologically intuitive directions.
While not requiring any labeled data, the performance of these methods is still far from that of supervised methods (Figure~\ref{fig:teaser}).


We continue in the vein of self-supervised methods and propose an improved tracker,
which we call \emph{Memory-Augmented Self-Supervised Tracker} (MAST).
Similar to previous self-supervised methods, 
our model performs tracking by learning a feature representation that enables robust pixel-wise correspondences between frames; 
it then propagates a given segmentation mask to subsequent frames based on the correspondences. 
We make three main contributions:
\emph{first},
we reassess the traditional choices used for self-supervised training and reconstruction loss 
by conducting thorough experiments to finally determine the optimal choices. 
\emph{Second}, to resolve the challenge of tracker drift (\ie~as the object changes appearance or becomes occluded,
each subsequent prediction becomes less accurate if propagated only from recent frames),
we further improve on existing methods by augmenting our architecture with a crucial memory component. 
We design a coarse-to-fine approach that is necessary to efficiently access the memory bank:
a two-step attention mechanism first coarsely searches for candidate windows,
and then computes fine-grained matching.
We conduct experiments to analyze our choice of memory frames,
showing that both short- \emph{and} long-term memory
are crucial for good performance.
\emph{Third}, 
we benchmark on large-scale video segmentation datasets and propose a new metric, \emph{i.e.}~generalizability,
with the goal of measuring the performance gap between tracking seen and unseen categories, 
which we believe better captures the real-world use-cases for category-agnostic tracking.

The result of the first two contributions 
is a self-supervised network that surpasses 
all existing approaches by a significant margin on DAVIS-2017 ($15\%$) and YouTube-VOS ($17\%$) benchmarks,
making it competitive with supervised methods \emph{for the first time}.
Our results show that a strong representation for tracking can be learned without using any semantic annotations,  
echoing the early-stage development of the human visual system.
Beyond significantly narrowing the gap with supervised methods on the existing metrics,
we also demonstrate the \emph{superiority} of self-supervised approaches over supervised methods on generalizability.
On the unseen categories of YouTube-VOS benchmark, 
we surpass PreMVOS~\cite{Luiten18}, 
the 2018 challenge winner algorithm trained on massive segmentation datasets.
Furthermore, when we analyze the drop in performance between seen and unseen categories,
we show that our method (along with other self-supervised methods) has a significantly smaller
\emph{generalization gap} than supervised methods.
These results show that contrary to the popular belief that self-supervised methods are not yet useful
due to their weaker performance, 
their greater generalization capability (due to not being at risk of overfitting to labels)
is actually a more desirable quality when being deployed in real-world settings,
where the domain gap can be significant.

%% file: sec/related.tex
\section{Related Work}
\par \noindent \textbf{Dense tracking~(\emph{aka.}~semi-supervised video segmentation)}
has typically been approached in one of two ways:
propagation-based or detection/segmentation-based.
The former approaches formulate the dense tracking task as a mask propagation problem from the first frame to the consecutive frames.
To leverage the temporal consistency between two adjacent frames, 
many propagation-based methods often try to establish dense correspondences with optical flow or metric learning~\cite{Khoreva17, Hu17, Hu18, Luiten18, Voigtlaender19}.
However, computing optical flow remains a challenging, yet unsolved problem. 
Our method relaxes the constraint of optical flow's one-to-one brightness constancy constraint and spatial smoothness,
allowing each query pixel to potentially build correspondence with multiple reference pixels. 
On the other hand, 
detection/segmentation-based approaches address the tracking task with sophisticated detection or segmentation networks, 
but since these models are usually not class-agnostic during training,
they often have to be fine-tuned on the first frame of the target video during inference~\cite{Caelles17,Maninis18, Luiten18},
whereas our method requires no fine-tuning. 
\\[-8pt]

\par \noindent \textbf{Self-supervised learning on videos} 
has generated fruitful research in recent years. 
Due to the abundance of online data~\cite{Agrawal15,Denton17,Fernando17, Gan18,Jakab18,Jia16, Lee17, Misra16,Wang15,Wiles18,Wiles18a, Isola15,Jayaraman16,Jayaraman15,Jing18,Kim18,Vondrick18},
various ideas have been explored to learn representations by exploiting the spatio-temporal information in videos.
~\cite{Misra16,Fernando17,Wei18} 
exploit spatio-temporal ordering for learning video representations.
Recently, Han~\etal~\cite{Han19} learn strong video representations for action recognition by self-supervised contrastive learning on raw videos. 
Of more relevance, \cite{Vondrick18, Lai19} have recently leveraged the natural temporal coherency of color in videos, 
to train a network for tracking and correspondence related tasks. We discuss these works in more detail in Section~\ref{sec:background}.
In this work, we propose to augment the self-supervised tracking algorithms with a differentiable memory module.
We also rectify some flaws in their training process.
\\[-8pt]

\par \noindent \textbf{Memory-augmented models}
refer to the computational architecture that has access to a memory repository for prediction.
Such models typically involve an internal memory implicitly updated in a recurrent process, 
\eg LSTM~\cite{Hochreiter97} and GRU~\cite{chung2014empirical}, 
or an explicit memory that can be read or written with an attention-based procedure~\cite{Xu2015,Vaswani17,Bahdanau2015,Wang2018,Devlin18,Kumar2016,Graves2014,Sukhbaatar2015}. 
Memory models have been used for many applications, 
including reading comprehension~\cite{Sukhbaatar2015}, 
summarization~\cite{See2017}, tracking\cite{Zhu17}, video understanding\cite{Wu19}, and
image and video captioning~\cite{Xu2015,Yao2015}. 
In dense visual tracking, 
the popular memory-augmented models treat key frames as memory~\cite{Seoung19}, 
and use attention mechanisms to read from the memory.
Despite its effectiveness, 
the process of computing attention either does not scale to multiple frames or is unable to process high-resolution frames, 
due to the computational bottleneck in hardware, \eg~physical memory.
In this work,
we propose a scalable way to process high-resolution information in a coarse-to-fine manner.
The model enables dynamic localization of salient regions, 
and fine-grained processing is only required for a small fraction of the memory bank.

%% file: sec/method.tex
\section{Method}
The proposed dense tracking system, 
MAST (Memory-Augmented Self-Supervised Tracker), 
is a conceptually simple model for dense tracking that can be trained with self-supervised learning,
\ie \textbf{zero manual annotation} is required during training,
and an object mask is only required for the first frame during inference.
In Section~\ref{sec:background},
we provide relevant background of previous self-supervised dense tracking algorithms,
and terminologies that will be used  in later sections.
Next, in Section~\ref{sec:training_signal}, 
we pinpoint weaknesses in these works and propose improvements to the training signals.
Finally, in Section~\ref{sec:memory}, we propose memory augmentation as an extension to existing self-supervised trackers.

\subsection{Background}
\label{sec:background}
In this section,
we review previous papers that are closely related to this work~\cite{Vondrick18, Lai19}.
In general, 
the goal of self-supervised tracking is to learn feature representations that enable robust correspondence matching.
During training, 
a proxy task is posed as reconstructing a target frame~($I_t$) by linearly combining pixels from a reference frame~($I_{t-1}$),
with the weights measuring the strength of correspondence between pixels.

Specifically, 
a triplet~($\{Q_t, K_t, V_t\}$) exists for each input frame $I_t$, 
referring to \emph{Query}, \emph{Key}, and \emph{Value}.
In order to reconstruct a pixel $i$ in the $t$-th frame~($\hat{I^i_t}$),
an \textit{Attention} mechanism is used for \emph{copying} pixels from a subset of previous frames in the original sequence.
This procedure is formalized as:
\begin{align} 
\label{eq:attention}
  \hat{I^i_t} = \sum _{j} A^{ij}_{t} V^j_{t-1} \\
\label{eq:affinity}
  A^{ij}_t =  \frac{exp\langle\ Q_t^i,K_{t-1}^j\rangle}{\sum_p exp\langle\ Q_t^i,K_{t-1}^p\rangle}
\end{align}

\noindent where $\langle\cdot,\cdot\rangle$ refers to the dot product between two vectors, 
query~(Q) and key~(K) are feature representations computed by 
passing the target frame $I_t$ to a Siamese ConvNet $\Phi(\cdot; \theta)$, 
\ie $Q_t = K_t = \Phi(I_t; \theta)$,
$A_t$ is the affinity matrix representing the feature similarity between pixel $I^i_t$ and $I^j_{t-1}$,
value~(V) is the raw reference frame~($I_{t-1}$) during the training stage, 
and instance segmentation mask during inference,
achieving reconstruction or dense tracking respectively.

A key element in self-supervised learning is to set the proper \emph{information bottleneck},
or the choice of what input information to withhold for learning the desired feature representation 
and avoiding trivial solutions.
For example, in the reconstruction-by-copying task,
an obvious shortcut is that the pixel in $I_t$ can learn to match any pixel in $I_{t-1}$ with the \emph{exact same color}, 
yet not necessarily correspond to the same object.
To circumvent such learning shortcuts,
Vondrick \etal~\cite{Vondrick18} intentionally drop the color information from the input frames.
Lai and Xie~\cite{Lai19} further show that a simple channel dropout can be more effective.

\subsection{Improved Reconstruction Objective} 
\label{sec:training_signal}
In this section,
we reassess the choices made in previous self-supervised dense tracking works and provide intuition for our optimal choices,
which we empirically support in Section \ref{sec:experiments}.
\vspace{-5pt}
\subsubsection{Decorrelated Color Space}
\label{sec:color}
\begin{figure}[!h]
  \centering
  \includegraphics[width=.45\textwidth]{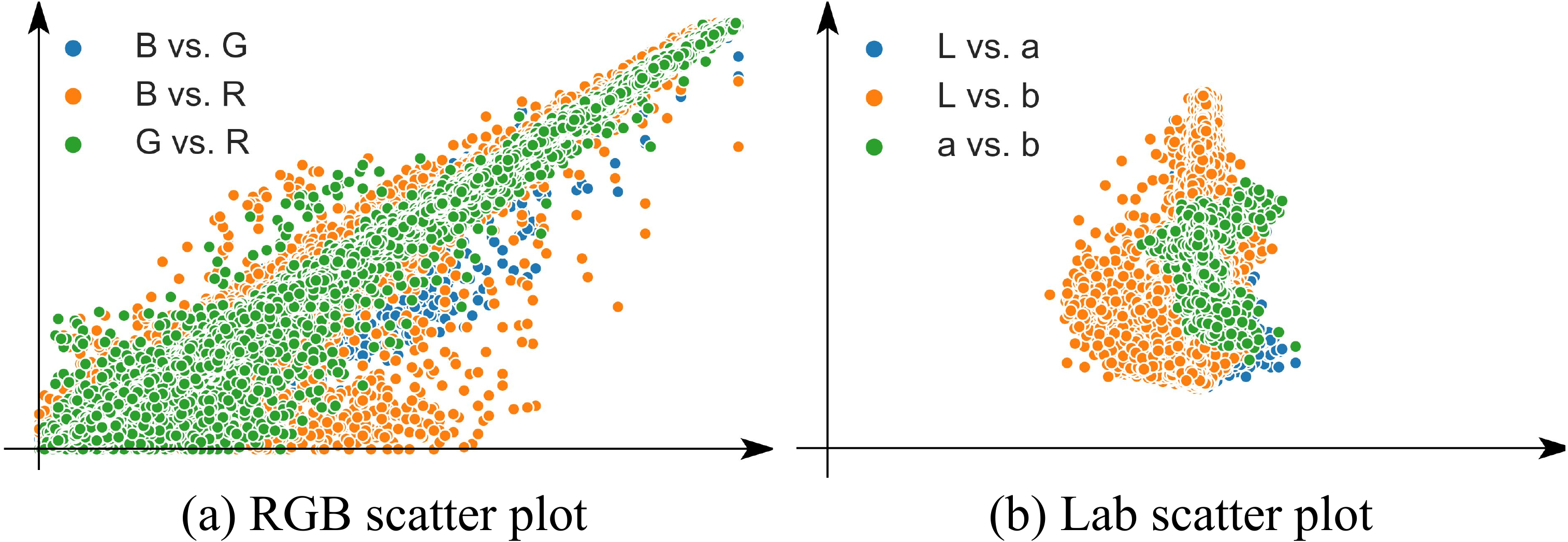}
  \vspace{-2pt}
  \caption{Correlation between channels of RGB and $Lab$ colorspace. 
We randomly take $100,000$ pixels from 65 frames in a sequence~(snowboard) in the DAVIS dataset and plot the relative relationships between RGB channels. This phenomena generally holds for all natural images~\cite{Reinhard01}, 
due to the fact that all of the channels include a representation of brightness.
Values are normalized for visualization purposes.}
  \label{fig:color}
\end{figure}

Extensive experiments in the human visual system have shown that colors can be seen as combinations of the primary colors,
namely red (R), green (G) and blue (B).
For this reason, 
most of the cameras and emissive color displays represent pixels as a triplet of intensities:~$(R,G,B) \in \mathcal{R}^3$.
However,
a disadvantage of the RGB representation is that the channels tend to be extremely correlated~\cite{Reinhard01},
as shown in Figure~\ref{fig:color}.
In this case, 
the channel dropout proposed in~\cite{Lai19} is unlikely to behave as an effective information bottleneck,
since the dropped channel can almost always be determined by one of the remaining channels.

To remedy this limitation,
we hypothesize that dropout in the decorrelated representations~(\eg~$Lab$) 
would force the model to learn invariances suitable for self-supervised dense tracking;
\ie if the model cannot predict the missing channel from the observed channels,
it is forced to learn a more robust representation rather than relying on local color information.

\vspace{-5pt}
\subsubsection{Classification vs. Regression}
\label{sec:loss}
In the recent literature on colorization and generative models~\cite{Zhang16_Color,Oord16},
colors were quantized into discrete classes and treated as a multinomial distribution,
since generating images or predicting colors from grayscale images is usually a non-deterministic problem; 
\eg the color of a car can reasonably be red or white.
However, this convention is suboptimal for self-supervised learning of correspondences,
as we are not trying to generate colors for each pixel,
but rather, estimate a precise relocation of pixels in the reference frames.
More importantly,
quantizing the colors leads to an information loss that can be crucial for learning high-quality correspondences.

We conjecture that directly optimizing a regression loss between the reconstructed frame~($\hat{I_{t}}$) and real frame~($I_t$) will provide more discriminative training signals.
In this work, the objective $\mathcal{L}$ is defined as the Huber Loss:
\begin{align}
\mathcal{L} = \frac{1}{n} \sum_{i} z_{i}
\end{align}
\vspace{-.3cm}
\noindent where
\begin{align}
 z_{i} =
        \begin{cases}
        0.5 (\hat{\textbf{I}}^i_t - \textbf{I}^i_t)^2, & \text{if } |\hat{\textbf{I}}^i_t - \textbf{I}^i_t| < 1 \\
        |\hat{\textbf{I}}^i_t - \textbf{I}^i_t| - 0.5, & \text{otherwise }
        \end{cases}
\end{align}
where $\hat{\textbf{I}}^i_t \in \mathcal{R}^3$ refers to RGB or \emph{Lab}, 
normalized to the range [-1,1] in the reconstructed frame that is copied from pixels in the reference frame~$I_{t-1}$,
and $I_t$ is the real frame at time point $t$.


\begin{figure*}[!htb]
  \centering
  \includegraphics[width=.98\textwidth]{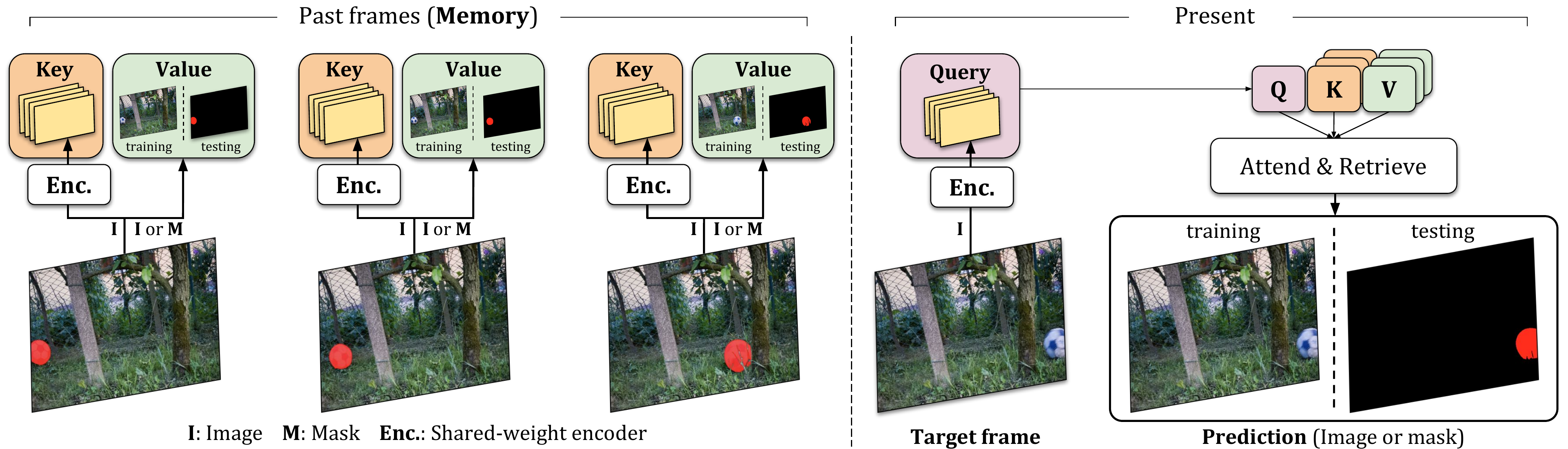}
  \vspace{-3pt}
  \caption{Structure of MAST. 
  The current frame is used to compute \textbf{query} to attend and retrieve from memory (\textbf{key} \& \textbf{value}). 
  During training, we use raw video frame as \textbf{value} for self-supervision. 
  Once the encoder is trained, we use instance mask as \textbf{value}. See Section~\ref{sec:memory} for details.}
  \vspace{-5pt}
  \label{fig:scheme}
\end{figure*}

\vspace{5pt}
\subsection{Memory-Augmented Tracking}
\label{sec:memory}
So far we have discussed the straightforward attention-based mechanism for propagating
a mask from a single previous frame.
However, as predictions are made recursively,
errors caused by object occlusion and disocclusion tend to accumulate 
and eventually degrade the subsequent predictions.
To resolve this issue, we propose an attention-based tracker that efficiently makes use of \emph{multiple} reference frames. 

\subsubsection{Multi-frame tracker}
An overview of our tracking model is shown in Figure~\ref{fig:scheme}.
To summarize the tracking process: 
given the present frame and multiple past frames (memory bank) as input,
we first compute the query~(Q) for the present frame and keys~(K) for all frames in memory.
Here, we follow the general procedure in previous works as described in Section~\ref{sec:background},
where K and Q are computed from a shared-weight feature extractor and
V is equal to the input frame (during training) or object mask (during testing).
The computed affinity between Q and all the keys~(K) in memory is 
then used to make a prediction for each query pixel depending on V.
Note we don't put any weights on the reference frames,
as this should be encoded in the affinity matrix 
(\emph{e.g.}~when a target and reference frame are dis-similar,
the corresponding similarity value will be naturally low;
thus the reference label will have less contribution to the labeling of a target pixel).

The decision of which pixels to include in K is crucial for good performance.
Including all pixels previously seen is far too computationally expensive due to the quadratic explosion of the affinity matrix (\eg
the network of~\cite{Lai19} produces affinity matrices with more than 1 billion elements for 480p videos).
To reduce computation, 
\cite{Lai19} exploit temporal smoothness in videos and apply restricted attention, 
only computing the affinity with pixels in a ROI around the query pixel location.
However, the temporal smoothness assumption holds only for temporally close frames.

To efficiently process temporally distant frames, we propose a two-step attention mechanism.
The first stage involves coarse pixel-matching with the frames in the memory bank to
determine which ROIs are likely to contain good matches with the query pixel.
In the second stage, we extract the ROIs and compute fine-grained pixel matching,
as described in Section~\ref{sec:background}.
Overall, the process can be summarized in Algorithm~\ref{algorithm}.
\setlength{\textfloatsep}{2pt}
\begin{algorithm}[!htb]
\small
  \caption{MAST} 
  \begin{algorithmic}[1]
    \State Choose $m$ reference frames $Q_1, Q_2, ... Q_m$
    \State Localize ROI $R_1, R_2, ... R_m$ according to 3.3.2 (Eq. 5 and 6) for each of the reference frames 
    \State Compute similarity matrix $A^{ij}_t = \langle Q^j,R_t^i\rangle $  between target frame $Q$ and each ROI.
    \State Output: pixel's label is determined by aggregating the labels of the ROI pixels (weighted by its affinity score).
  \end{algorithmic} 
\label{algorithm}
\end{algorithm}

\vspace{-15pt}
\subsubsection{ROI Localization}
\label{sec:localization}
\begin{figure}[H]
  \centering
  \includegraphics[width=.42\textwidth]{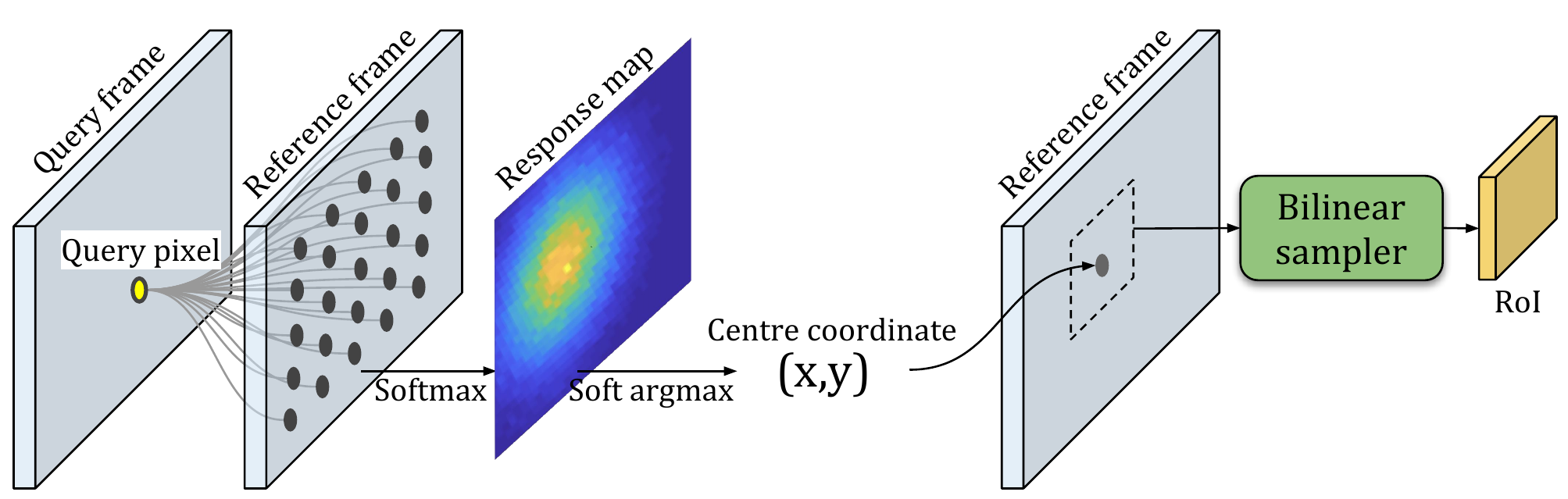}
  \vspace{-3pt}
  \caption{ROI Localization.}
  \label{fig:ROI}
\end{figure}

The goal of ROI localization is to estimate the candidate windows non-locally from memory banks. As shown in Figure~\ref{fig:ROI}, this can be achieved through restricted attention with varying dilation rate.

Intuitively, for short-term memory (temporally close frames),
dilation is not required since spatial-temporal coherence naturally exists in videos;
thus ROI localization becomes restricted attention~(similar to \cite{Lai19}).
However, for long-term memory,
we aim to account for the fact that objects can potentially appear anywhere in the reference frames.
We unify both scenarios into a single framework for learning ROI localization.

Formally, for the query pixel $i$ in $I_t$, to localize the ROI from frame~($I_{t-N}$),
we first compute in parallel $H^i_{t-N, x,y}$, the similarity heatmap between $i$ and all candidate pixels in the dilated window:
\begin{align}
H^i_{t-N, x,y} = softmax(Q^i_t \cdot im2col(K^i_{t-N}, \gamma_{t-N}))
\end{align}
where $\gamma_{t-N}$ refers to the dilation rate for window sampling in frame $I_{t-N}$,
and $im2col$ refers to an operation that transforms the input feature map into a matrix based on dilation rate. 
Specifically, in our experiments, the dilation rate is proportional to the temporal distance between the present frame and the past frames in the memory bank, \ie~$\gamma_{t-N} \propto N$. 
We use $\gamma_{t-N} = \left \lceil (t-N)  / 15 \right \rceil$.

The center coordinates for ROIs can be then computed via a \textit{soft-argmax} operation:
\begin{align}
P^i_{x,y} = \sum_{x,y} H^i_{x,y} * C 
\end{align}
where $P^i_{x,y}$ is the estimated center location of the candidate window in frame $I_{t-N}$ for query pixel $I^i_t$,
and $C$ refers to the grid coordinates~($x,y$) corresponding to the pixels in the window from $im2col$.
The resampled Key~($\hat{K}^i_{t-N}$) for pixel $I^i_t$ can be extracted with a bilinear sampler~\cite{Jaderberg2015}.
With all the candidate Keys dynamically sampled from different reference frames of the memory bank, 
we compute fine-grained matching scores only with these localized Keys,
resembling a restricted attention in a non-local manner.
Therefore, with the proposed design, 
the memory-augmented model can efficiently access high-resolution information for correspondence matching, 
without incurring large physical memory costs.  

%% file: sec/imp.tex
\section{Implementation Details}
\begin{figure}[t]
  \centering
  \includegraphics[width=.48\textwidth]{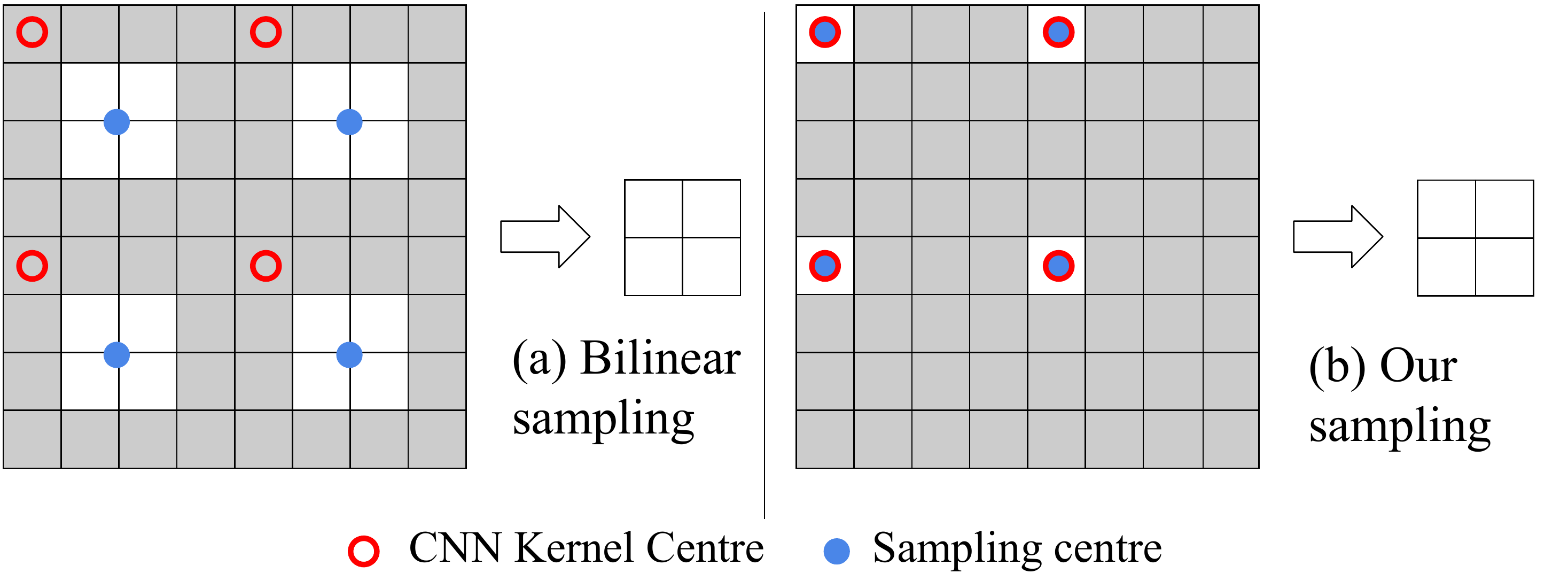}
  \vspace{-15pt}
  \caption{Image Feature Alignment. We fix the misalignment between strided convolution layers (with kernel centered at red circle) and images from na\"ive bilinear downsampling (with kernel centered at blue dot) by sampling directly at the strided convolution centers.}
  \label{fig:align}
\end{figure}

\noindent\textbf{Training: }
For fair comparison,
we adopt as our feature encoder the same architecture~(ResNet18) as~\cite{Lai19} in all experiments~(as shown in Supplementary Material).
The network produces feature embeddings with a spatial resolution $1/4$ of the original image.
The model is trained in a completely self-supervised manner,
meaning the model is initialized with random weights, 
and we do \emph{not} use any information other than raw video sequences.
We report main results on two training datasets: OxUvA~\cite{Valmadre18} and YouTube-VOS (both raw videos only). 
We report the first for fair comparison with the state-of-the-art method~\cite{Lai19} and the second for maximum performance. 
As pre-processing, we resize all frames to $256\times 256\times3$. 
In all of our experiments, 
we use $I_0$,  $I_5$~(only if the index for the current frame is larger than $5$) as long term memory, 
and {$I_{t-5}$, $I_{t-3}$, $I_{t-1}$} as short term memory. 
Empirically, we find the choice of frame number has small impact on performance, 
but using both long and short term memory is essential~(See appendix). \\[3pt]
During training, 
we first pretrain the network with a pair of input frames, 
\ie~one reference frame and one target frame are fed as inputs.
One of the color channels is randomly dropped with probability $p=0.5$.
We train our model end-to-end using a batch size of $24$ for 1M iterations with the Adam optimizer. 
The initial learning rate is set to $1e$-3, and halved after 0.4M, 0.6M and 0.8M iterations. 
We then finetune the model using multiple reference frames (our full memory-augmented model) 
with a small learning rate of $2e$-5 for another 1M iterations.
As discussed in Section~\ref{sec:loss}, 
the model is trained with a photometric loss between the reconstruction and the true frame. \\[3pt]
\noindent\textbf{Inference: }
We use the trained feature encoder to compute the
affinity matrix between pixels in the target frame and those in the reference frames. 
The affinity matrix is then used to propagate the desired pixel-level entities,
such as instance masks in the dense tracking case, 
as described in Algorithm~\ref{algorithm}.  \\[3pt]
\noindent\textbf{Image Feature Alignment: }
Due to memory constraints, 
the supervision signals in previous methods were all defined on bilinearly downsampled images.
As shown in Figure~\ref{fig:align}(a), this introduces a misalignment between strided convolution layers 
and images from na\"ive bilinear downsampling. 
We handle this spatial misalignment between feature embedding and image by directly sampling at the strided convolution centers.
This seemingly minor change actually brings significant improvement to the downstream tracking task (Table~\ref{tab:abl_align}).


%% file: sec/exp.tex
\section{Experiments}
\label{sec:experiments}
\input{./tables/davis-2017.tex}
We benchmark our model on two public benchmarks:
DAVIS-2017~\cite{Pont-Tuset17}  and the current largest video segmentation dataset, YouTube-VOS~\cite{Xu18}.
The former contains 150 HD videos with over 30K manual instance segmentations,
and the latter has over 4000 HD videos of 90 semantic categories, totalling over 190k instance segmentations.
For both datasets,
we benchmark the proposed self-supervised learning architecture~(MAST) on the official semi-supervised video segmentation setting~(\emph{aka.}~dense tracking),
where a ground truth instance segmentation mask is given for the first frame, 
and the objective is to propagate the mask to subsequent frames.
In Section~\ref{sec:davis},  
we report performance of our full model and several ablated models on the DAVIS benchmark.
Next, in Section~\ref{sec:youtube_vos}, we analyze the \emph{generalizability} of our model by benchmarking on the large-scale YouTube-VOS dataset.

\begin{figure*}[!htb]
  \centering
  \includegraphics[width=\textwidth]{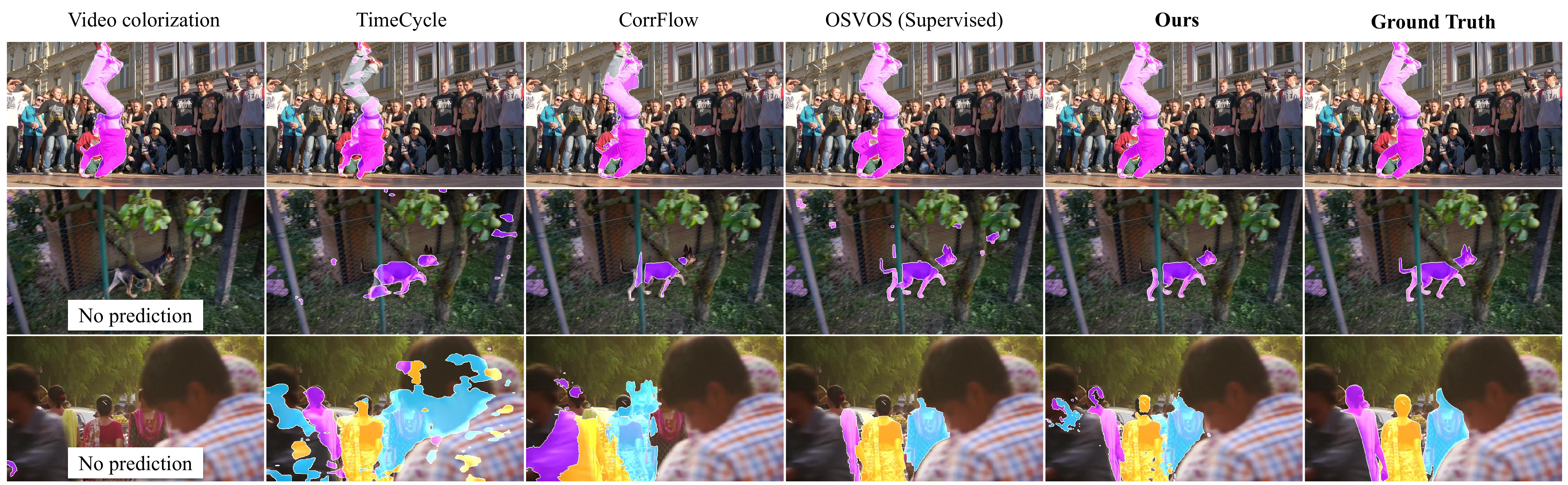}
  \vspace{-10pt}
  \caption{Our method vs. previous self-supervised methods. 
  Other methods show systematic errors in handling occlusions. Row 1: The dancer undergoes large self-occlusion. 
  Row 2: The dog is repeatedly occluded by poles. 
  Row 3: Three women reappear after being occluded by the man in the foreground.
  }
    \vspace{-1.5em}
  \label{fig:qualitative_2}
\end{figure*}
\vspace{3pt}
\par \noindent \textbf{Standard evaluation metrics.} 
We use region similarity~($\mathcal{J}$) and contour accuracy~($\mathcal{F}$) to evaluate the tracked instance masks~\cite{Perazzi16}.

\vspace{3pt}
\par \noindent \textbf{Generalizability metrics} 
To demonstrate the generalizability of tracking algorithms in category-agnostic scenarios,
\ie~the categories in training set and testing set are disjoint,
YouTube-VOS also explicitly benchmarks the performances on unseen categories.
We therefore evaluate a \emph{generalization gap} in Section~\ref{sec:gen}, 
which is defined as the average performance difference between seen and unseen object classes:
\begin{align}
\text{Gen. Gap} = \frac{(\mathcal{J}_{seen} - \mathcal{J}_{unseen}) + (\mathcal{F}_{seen}  - \mathcal{F}_{unseen})}{2}
\end{align}
Note, the proposed metric aims to explicitly penalize the case where the performance on seen outperforms unseen by large margins,
while at the same time provide a reward when the performance on unseen categories is higher than on seen ones.

\subsection{Video Segmentation on DAVIS-2017}
\label{sec:davis}
\subsubsection{Main results}
In Table~\ref{table:davis_1}, 
we compare MAST with previous approaches on the DAVIS-2017 benchmark. 
Two phenomena can be observed:
\emph{first},
our proposed model clearly outperforms all other self-supervised methods,
surpassing previous state-of-the-art CorrFlow by a significant margin~($65.5$ vs $50.3$ on $\mathcal{J}$\&$\mathcal{F}$).
\emph{Second}, 
despite using only ResNet18 as the feature encoder,
our model trained with self-supervised learning can still surpass supervised approaches that use heavier architectures.

\vspace{-10pt}
\subsubsection{Ablation Studies}
\label{exp:ablation}
To examine the effects of different components, 
we conduct a series of ablation studies by removing one component at a time.
All models are trained on OxUvA~(except for the analysis on different datasets), 
and evaluated on DAVIS-2017 semi-supervised video segmentation~(\emph{aka.}~dense tracking) \emph{without} any finetuning.\\[-8pt]
\input{tables/ablation.tex}

\par \noindent  \textbf{Choice of color spaces.}
As shown in Table~\ref{tab:abl_color}, 
we perform different experiments with input frames transformed into different color spaces, \eg~RGB, $Lab$ or HSV.
We find that the MAST model trained with $Lab$ color space always outperforms the other color spaces,
validating our conjecture that dropout in a decorrelated color space leads to better feature representations for self-supervised dense tracking, as explained in Section~\ref{sec:color}.
Additionally, 
we compare our default setting with a model trained with cross-color space matching task~(shown in Table~\ref{tab:abl_cross_color}).
That means to use a different color space for the input and the training objective, 
\eg~input frames are in RGB, and loss function is defined in $Lab$ color space.
Interestingly, the performance drops significantly, 
we hypothesis this can attribute to the fact that all RGB channels include a representation of brightness,
making it highly correlate to the luminance in $Lab$, 
therefore acting as a weak information bottleneck.\\[-8pt]
\par \noindent  \textbf{Loss functions.}
As a variation of our training procedure, 
we experiment with different loss functions: 
cross entropy loss on the quantized colors, and photometric loss with Huber loss.
As shown in Table~\ref{tab:abl_color}, 
regression with real-valued photometric loss surpasses classification significantly,
validating our conjecture that the information loss during color quantization results in inferior representations for self-supervised tracking~(as explained in Section~\ref{sec:training_signal}),
due to less discriminative training signals.\\[-8pt]
\par \noindent  \textbf{Image feature alignment.}
To evaluate the alignment module proposed for aligning features with the original image, 
we compare it to direct bilinear image downsampling used by CorrFlow~\cite{Lai19}. 
The result in Table~\ref{tab:abl_align} shows that our approach achieves about $2.2\%$ higher performance.\\[-8pt]

\par \noindent  \textbf{Dynamic memory by exploiting more frames.}
We compare our default network with variants that have only short term memory or long term memory. 
Results are shown in Table~\ref{tab:abl_mem}.
While both short term memory and long term memory alone can make reasonable predictions,
the combined model achieves the highest performance.
The qualitative predictions~(Figures~\ref{fig:qualitative} and~\ref{fig:qualitative_2}) also confirm that the improvements come from reduced tracker drift.
For instance, when severe occlusion occurs, 
our model is able to attend and retrieve high-resolution information from frames that are temporally distant. 

\subsection{Youtube Video Object Segmentation}
\label{sec:youtube_vos}
We also evaluate the MAST model on the Youtube-VOS validation split~(474 videos with 91 object categories).
As no other self-supervised methods have been tested on the benchmark, 
we directly compare our results with supervised methods. 
As shown in Table~\ref{table:youtube_1}, 
our method outperforms the other self-supervised learning approaches by a significant margin~($64.2$ vs.~$46.6$), 
and even achieves comparable performance to many heavily supervised methods. 

\input{tables/youtube-vos.tex}
\subsubsection{Generalizability}
\label{sec:gen}
As another metric for evaluating category-agonostic tracking, 
the YouTube-VOS dataset conveniently has separate measures for seen and unseen object categories. 
We can therefore estimate testing performance on out-of-distribution samples to gauge the model's generalizability to more challenging, unseen, real-world scenarios. 
As seen from the last two columns, we rank second amongst all algorithms in unseen objects. 
In these unseen classes, 
we are even $3.9\%$ higher than the DAVIS 2018 and YouTube-VOS 2018 video segmentation challenge winner, 
PreMVOS\cite{Luiten18}, 
a complex algorithm trained with multiple large manually labeled datasets.
For fair comparison, 
we train our model only on the YouTube-VOS training set. We also re-train two most relevant self-supervised methods in the same manner as baselines. 
Even learning from only a subset of all classes, 
our model generalizes well to unseen classes,
with a generalization gap (\ie the performance difference between seen and unseen objects) near zero ($0.4$). 
This gap is much smaller than any of the baselines ($\text{avg}=11.5$), 
suggesting a unique advantage to most other algorithms trained with labels. 

By training on large amounts of unlabeled videos, 
we learn an effective tracking representation without the need for any human annotations. 
This means that the learned network is not limited to a specific set of object categories (\ie those in the training set),
but is more likely to be a ``universal feature representation'' for tracking.
Indeed, the only supervised algorithm that is comparable to our method in generalizability is OSVOS (2.7 \textit{vs.} 0.4). 
However, OSVOS uses the first image from the testing sequence to perform costly domain adaptation, \eg~one-shot fine-tuning. 
In contrast, our algorithm requires no fine-tuning, which further demonstrates its zero-shot generalization capability. 

Note our model also has a smaller generalization gap compared to other self-supervised methods as well. 
This further attests to the robustness of its learned features, 
suggesting that our improved reconstruction objective is highly effective in capturing general features. 


%% file: tables/davis-2017.tex

\begin{table*}[!htb]
\centering
\footnotesize\addtolength{\tabcolsep}{-5pt}
\begin{tabular*}{\textwidth}{c @{\extracolsep{\fill}} ccccccccc}
\toprule
Method & Backbone & Supervised & Dataset (Size) & $\mathcal{J}$\&$\mathcal{F}$(Mean) $\uparrow$ & $\mathcal{J}$(Mean) $\uparrow$ & $\mathcal{J}$(Recall) $\uparrow$ & $\mathcal{F}$(Mean) $\uparrow$ & $\mathcal{F}$(Recall) $\uparrow$ \\ \midrule

Vid. Color.~\cite{Vondrick18}  & ResNet-18 & \xmark   & Kinetics (800 hours) & 34.0 & 34.6 & 34.1 & 32.7 & 26.8\\
CycleTime$^\dagger$~\cite{Wang19} & ResNet-50  & \xmark   & VLOG (344 hours) & 48.7 & 46.4 & 50.0 & 50.0 & 48.0  \\
CorrFlow$^\dagger$~\cite{Lai19} & ResNet-18  & \xmark   & OxUvA (14 hours) & {50.3} & {48.4} & {53.2} & {52.2} & {56.0} \\
UVC$^\star$~\cite{Li19} & ResNet-18  & \xmark   & Kinetics (800 hours) & {59.5} & {57.7} & {68.3} & {61.3} & {69.8} \\
\textbf{MAST (Ours)} & ResNet-18 & \xmark   & OxUvA (14 hours)  & \textbf{63.7} & \textbf{61.2} & \textbf{73.2} & \textbf{66.3} & \textbf{78.3} \\
\textbf{MAST (Ours)} & ResNet-18 & \xmark   & YT-VOS (5.58 hours)  & \textbf{65.5} & \textbf{63.3} & \textbf{73.2} & \textbf{67.6} & \textbf{77.7} \\
\midrule
ImageNet~\cite{He16} & ResNet-50 & \cmark  & I (1.28M, 0) & 49.7 & 50.3 & - & 49.0 & - \\
OSMN~\cite{Yang18} & VGG-16 & \cmark  & ICD (1.28M, 227k) & 54.8 & 52.5 & 60.9& 57.1 & 66.1 \\
SiamMask~\cite{Wang19a} & ResNet-50 & \cmark  & IVCY (1.28M,  2.7M) & 56.4 & 54.3 & 62.8 & 58.5 & 67.5\\
OSVOS~\cite{Caelles17}   &  VGG-16 & \cmark  & ID (1.28M, 10k) & 60.3 & 56.6 & 63.8 & 63.9 & 73.8     \\
OnAVOS~\cite{Voigtlaender17}  & ResNet-38  & \cmark  & ICPD (1.28M, 517k) & 65.4 & 61.6 & 67.4 & 69.1 & 75.4     \\
OSVOS-S~\cite{Maninis18}& VGG-16 & \cmark  & IPD (1.28M, 17k) & 68.0 & 64.7 & 74.2 & 71.3 & 80.7     \\
FEELVOS~\cite{Voigtlaender19} & Xception-65 &\cmark  & ICDY (1.28M, 663k) & 71.5 &69.1 &79.1 & 74.0 & 83.8     \\
PReMVOS~\cite{Luiten18} & ResNet-101 &\cmark  & ICDPM (1.28M, 527k) & 77.8 & 73.9 & 83.1 & 81.8 & 88.9     \\
STM~\cite{Seoung19} & ResNet-50 &\cmark  & IDY (1.28M, 164k) & 81.8 & 79.2 & - & 84.3 & -     \\
\bottomrule
\end{tabular*}
\vspace{-8pt}
\caption{Video segmentation results on DAVIS-2017 validation set. 
Dataset notations:
I=ImageNet, 
V = ImageNet-VID,
C=COCO, D=DAVIS, 
M=Mapillary,
P=PASCAL-VOC
Y=YouTube-VOS. For size of datasets, we report (length of \textit{raw videos}) for self-supervised methods and (\#image-level annotations, \#pixel-level annotations) for supervised methods. 
$^\star$ denotes concurrent work. $^\dagger$ denotes highest results reported after original publication. Higher values are better. }
\label{table:davis_1}
\end{table*}

%% file: tables/ablation.tex

\begin{table*}[t]
\centering
\footnotesize\addtolength{\tabcolsep}{-1pt}
\vspace{-5pt}
\begin{minipage}{.3\textwidth}
\centering
\begin{tabular}{cccc}

\toprule
Colors.  & Loss & $\mathcal{J}$(Mean)  & $\mathcal{F}$(Mean) \\
\midrule
\multirow{2}{*}{RGB}   &  Cls.  & 42.5  & 45.3       \\
   &  Reg.  & 52.7  & 57.1       \\
\midrule

 \multirow{2}{*}{HSV} &  Cls.    & 32.5 & 35.3  \\
  &  Reg.    & 54.3 & 58.6  \\

\midrule

\multirow{2}{*}{Lab}  &  Cls.   & 47.1 & 48.9 \\ 
 &  Reg.  & \textbf{61.2}  & \textbf{66.3}\\ 

        \bottomrule

\end{tabular}
\vspace{-8pt}
\caption{\textbf{Training colorspaces and loss}: Our final model trained with Lab colorspace with regression loss outperforms all other models on dense tracking task. Higher values are better. }
\label{tab:abl_color}
\end{minipage} \hfill
\begin{minipage}{.3\textwidth}
\centering
\vspace{-1pt}
\begin{tabular}{cccc}
\toprule
Input  &  Loss & $\mathcal{J}$(Mean)  & $\mathcal{F}$(Mean) \\
\midrule
Lab   & RGB     & 48.2  & 52.0       \\ 
RGB   & Lab     & 46.8  & 49.9       \\ 
\midrule
Lab   & Lab     & \textbf{61.2}  & \textbf{66.3}    \\
        \bottomrule

\end{tabular}
\vspace{25pt}
\caption{\textbf{Cross color space matching \textit{vs.} single color space}: Cross color space matching shows inferior results compared to single color space.}
\label{tab:abl_cross_color}
\end{minipage} \hfill
\begin{minipage}{.3\textwidth}
\centering
\begin{tabular}{ccc}
        \toprule
        I-F Align  & $\mathcal{J}$(Mean)  & $\mathcal{F}$(Mean)   \\ \midrule

        No        & 59.1  & 64.0        \\
        Yes       & \textbf{61.2}  & \textbf{66.3}      \\ 
        \midrule
         & $+2.1$ & $+2.3$  \\
                 \bottomrule

\end{tabular}
\vspace{28pt}
\caption{\textbf{Image-Feature alignment}: Using the improved Image-Feature alignment implementation improves the results. Higher values are better. }
\label{tab:abl_align}
\end{minipage} \hfill
\vspace{5pt}
\begin{minipage}{.3\textwidth}
\centering
    \vspace{5pt}

    \begin{tabular}{cccc}
        \toprule 
        Memory  & $\mathcal{J}$(Mean)  & $\mathcal{F}$(Mean)   \\ 

        \midrule

        Only long   &  44.6 & 48.7 \\
        Only short    & 57.3 & 61.8   \\
        \midrule
        
        Both      & \textbf{61.2}  & \textbf{66.3}  \\

         \bottomrule
  
    \end{tabular}
    \vspace{-5pt}
    \caption{\textbf{Memory length}: Removing either long term or short term memory results in a performance drop.
 }

\label{tab:abl_mem}
\vspace{10pt}
\end{minipage}\hfill
\begin{minipage}{.3\textwidth}
\centering
\vspace{6pt}
\begin{tabular}{ccc}

        \toprule
        Propagation  & $\mathcal{J}$(Mean)  & $\mathcal{F}$(Mean)   \\ \midrule

        Soft        & 57.0  & 61.7        \\
        Hard       & \textbf{61.2}  & \textbf{66.3}      \\ 
        \midrule
         & $+4.2$ & $+4.6$  \\
                 \bottomrule

\end{tabular}
\vspace{-4pt}
\caption{\textbf{Soft \textit{vs.} hard propagation}: \textit{Quantizing} class probability of each pixel (hard propagation) shows large gains over propagating probility distribution (soft propagation). }
\label{tab:abl_prop}
\end{minipage} \hfill
\begin{minipage}{.35\textwidth}
\centering
    \vspace{6pt}

    \begin{tabular}{ccc}
    \toprule
    Dataset  & $\mathcal{J}$(Mean)  & $\mathcal{F}$(Mean)  \\ 
    \midrule
    OxUvA       & 61.2  & 66.3    \\
    ImageNet VID         & 60.0 & 63.9    \\ 
    YouTube-VOS (w/o anno.)    & 63.3 & 67.6 \\
            \bottomrule

    \end{tabular}
    \vspace{0pt}
    \caption{\textbf{Training dataset}: All datasets provide reasonable performance, with O and Y slightly superior. We conjecture that our model gains from higher quality videos and larger object classes in these datasets. }
\label{tab:abl_dataset}
\end{minipage}
\vspace{-0.2cm}
\end{table*}

%% file: tables/youtube-vos.tex

\begin{table}[!htb]
\centering
\footnotesize\addtolength{\tabcolsep}{-2pt}
\begin{tabular}{c @{\extracolsep{\fill}} ccccccc}
\toprule

\multirow{2}{*}{Method} & \multirow{2}{*}{Sup.}  & \multirow{2}{*}{Overall $\uparrow$} & \multicolumn{2}{c}{Seen} & \multicolumn{2}{c}{Unseen} & 
\multirow{2}{*}{Gen. Gap $\downarrow$}\\ 
\cmidrule{4-5} \cmidrule{6-7} 
 &  & & $\mathcal{J}$ $\uparrow$ & $\mathcal{F}$ $\uparrow$ & $\mathcal{J}$ $\uparrow$ & $\mathcal{F}$ $\uparrow$  \\ 

 \midrule

Vid. Color.\cite{Vondrick18}$^\dagger$  &  \xmark & 38.9  & 43.1  & 38.6  & 36.6 & 37.4 & 3.9 \\
CorrFlow\cite{Lai19}  &    \xmark & 46.6   & 50.6   & 46.6 & 43.8  & 45.6 &  3.9 \\
\textbf{MAST (Ours)}   &  \xmark & \textbf{64.2}  & \textbf{63.9}  & \textbf{64.9}  & \textbf{\textcolor{blue}{60.3}} & \textbf{\textcolor{blue}{67.7}} &  \textbf{0.4}\\
\midrule

OSMN\cite{Yang18}   & \cmark &  51.2  & 60.0 & 60.1 & 40.6  & 44.0  &  17.75\\
MSK\cite{Khoreva16}    & \cmark &  53.1  & 59.9  & 59.5  & 45.0  & 47.9 &  13.25 \\
RGMP\cite{Oh18}  & \cmark &  53.8  & 59.5  & -  & 45.2  & - &  14.3\\
OnAVOS\cite{Voigtlaender17}  & \cmark &  55.2  & 60.1  & 62.7  & 46.6  & 51.4 &  12.4\\
RVOS\cite{Ventura_2019_CVPR}  & \cmark &  56.8  & 63.6  & 67.2  & 45.5  & 51.0 &  17.15\\
OSVOS\cite{Caelles17}  & \cmark &  58.8  & 59.8  & 60.5  & 54.2  & 60.7 &  2.7\\
S2S\cite{Xu18}  & \cmark &  64.4  & 71.0  & 70.0  & 55.5  & 61.2 &  12.15\\
PreMVOS\cite{Luiten18}  & \cmark &  66.9  & 71.4  & 75.9  & 56.5  & 63.7 &  13.55\\
STM\cite{Seoung19}  & \cmark &  79.4  & 79.7  & 84.2  & \textcolor{red}{72.8}  & \textcolor{red}{80.9} &  5.1\\

\bottomrule
\end{tabular}
\vspace{-5pt}
\caption{Video segmentation results on Youtube-VOS dataset. Higher values are better. According to the evaluation protocol of the benchmark, we report performance separated into ``seen'' and ``unseen'' classes (``Seen'' with respect to training set). $^\dagger$ indicates results based on our reimplementation. The first- and second-best results on the unseen category are highlighted in red and blue, respectively.}
\label{table:youtube_1}
\end{table}

%% file: supp_sec.tex
\clearpage
\begin{appendices}

\section{Network architecture}
In the same way as CorrFlow~\cite{Lai19}, we use a modified ResNet-18~\cite{He16} architecture. Details of the network are illustrated in Table~\ref{table:network-detail}.
\begin{table}[h!]
\centering
\footnotesize\addtolength{\tabcolsep}{5pt}

\begin{tabular}{ccc}

 \toprule
 Stage & Output & Configuration \\
 \hline
 0 & $H\times W$ & Input image  \\
  conv1 & $H/2\times W/2$ & $7 \!\times\! 7$, 64, stride 2  \\
  conv2 & $H/2\times W/2$ &
$\begin{bmatrix}
3\times 3, 64 
\\ 
3\times 3, 64 
\end{bmatrix} \times2$ \\
  conv3 & 
$H/4\times W/4$ & 
  $\begin{bmatrix}
3\times 3, 128 
\\ 
3\times 3, 128
\end{bmatrix} \times2$  \\
  conv4 & $H/4\times W/4$ &
 
  $\begin{bmatrix}
3\times 3, 256 
\\ 
3\times 3, 256
\end{bmatrix} \times2$  \\
  conv5 & $H/4\times W/4$ & 
  $\begin{bmatrix}
3\times 3, 256 
\\ 
3\times 3, 256
\end{bmatrix} \times2$  \\

\bottomrule
\end{tabular}
\caption{Network architecture. Residual Blocks are shown in brackets (a residually connected sequence of operations). See \cite{He16} for details.
\label{table:network-detail}}
\end{table}

\section{Optimal memory size}
\begin{figure}[H]
  \centering
  \includegraphics[width=.45\textwidth]{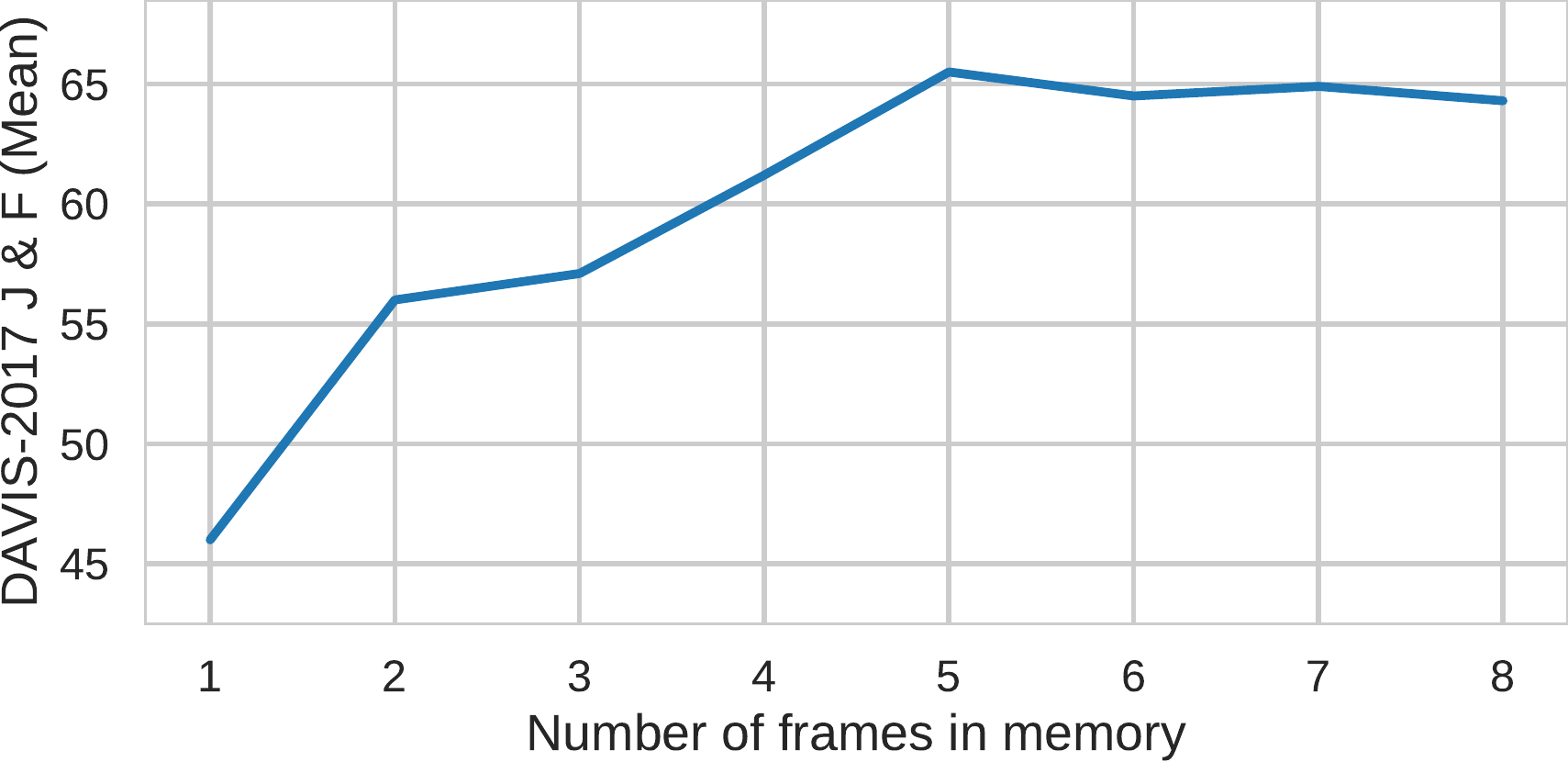}
  \vspace{-2pt}
  \caption{\textbf{Optimal memory size}: 
  Here, we test a changing memory size of $n+m$: $n$ short term memory and $m$ long term memory, where $n$ and $m$ grow alternatively. The performance of our model initially increases as the number of frames in memory grows, eventually plateauing at 5 frames. }
  \label{fig:mem_perf}
\end{figure}

In Figure~\ref{fig:mem_perf},
we explicitly show the effectiveness of increasing the number of reference frames in the memory bank, 
and confirm that a 5-frame memory is optimal for our task.

\section{Analysis by attributes}
\begin{figure*}[!htb]
  \centering
  \includegraphics[width=1.0\textwidth]
    {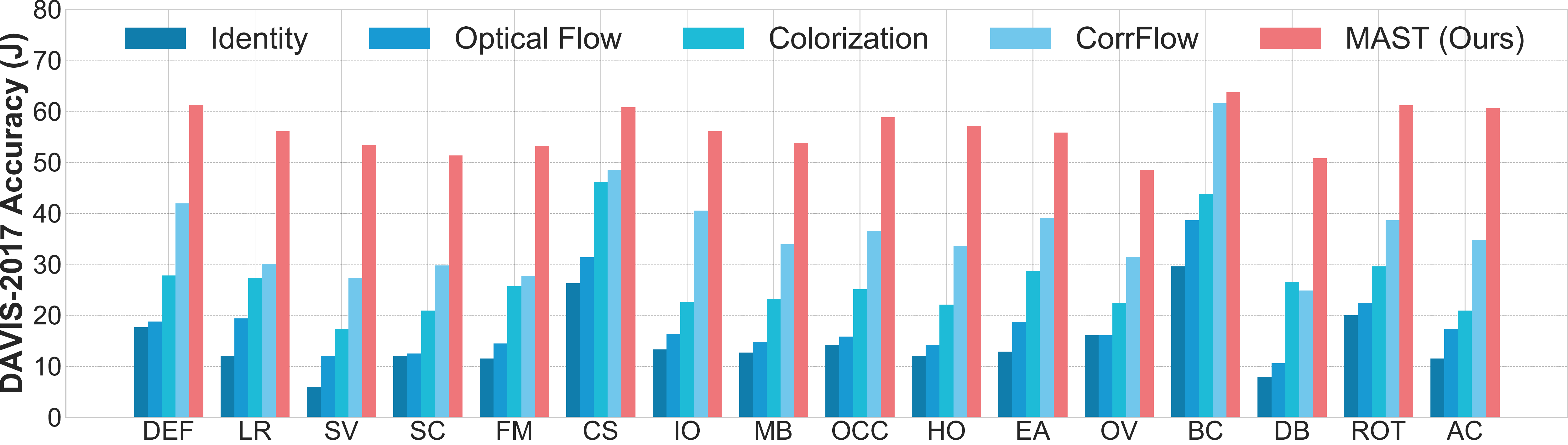}

  \captionof{figure}{\textbf{Accuracy broken down by attribute:} MAST outperforms previous self-supervised methods by a significant margin on all attributes, demonstrating the robustness of our model.}

  \label{fig:attributes}
\end{figure*}

We provide a more detailed accuracy list broken down by video attributes provided by the DAVIS benchmark~\cite{Perazzi16} (listed in Table~\ref{table:attributes}). 
The attributes illustrate the difficulties associated with each video sequence. Figure~\ref{fig:attributes} contains the accuracies categorized by attribute. 
Several trends emerge: \emph{first,} MAST outperforms all other self-supervised and unsupervised models by a large margin in all attributes. 
This shows that our model is robust to various challenges in dense tracking. 
\emph{Second,} MAST obtains significant gains on occlusion-related video sequences (\eg~OOC, OV), 
suggesting that memory-augmentation is a key enabler for high-quality tracking: 
retrieving occluded objects from previous frames is very difficult without memory augmentation. 
\emph{Third,} in videos involving background clutter, \ie background and foreground share similar colors, 
MAST obtains a relatively small improvement over previous methods. 
We conjecture this bottleneck could be caused by a shared photometric loss; 
thus a different loss type (\eg~based on texture consistency) could further improve the result.

\begin{table}[H]
\centering
\footnotesize\addtolength{\tabcolsep}{-1pt}
\begin{tabular}{cc|cc}
\toprule
ID  &  Description & ID  &  Description \\
\midrule
AC    &  Appearance Change   &   IO    &  Interacting Objects \\
BC    &  Background Clutter   &   LR    &  Low Resolution \\
CS    &  Camera-Shake   &   MB    &  Motion Blur  \\
DB    &  Dynamic Background   &   OCC    &  Occlusion  \\
DEF    &  Deformation   &   OV    &  Out-of-view  \\
EA    &  Edge Ambiguity   &   ROT    &  Rotation \\
FM    &  Fast-Motion   &   SC    &  Shape Complexity \\
HO    &  Heterogeneus Object  &   SV    &  Scale-Variation  \\
\bottomrule
\end{tabular}
\caption{List of video attributes provided in the DAVIS benchmark. We break down the validation accuracy according to the attribute list.}
\label{table:attributes}
\end{table}

\section{YouTube-VOS 2019 dataset}
We also evaluate MAST and two other self-supervised methods on YouTube-VOS 2019 validation dataset. The numerical results are reported in Table~\ref{table:youtube_19}. Augmenting on the 2018 version, the 2019 version contains more videos and object instances. We observe similar trend as reported in the main paper (\ie significant improvement and lower generalization gap).
\input{tables/youtube-vos-19.tex}

\section{More qualitative results}
\begin{figure*}[!htb]
  \centering
  \includegraphics[width=\textwidth]{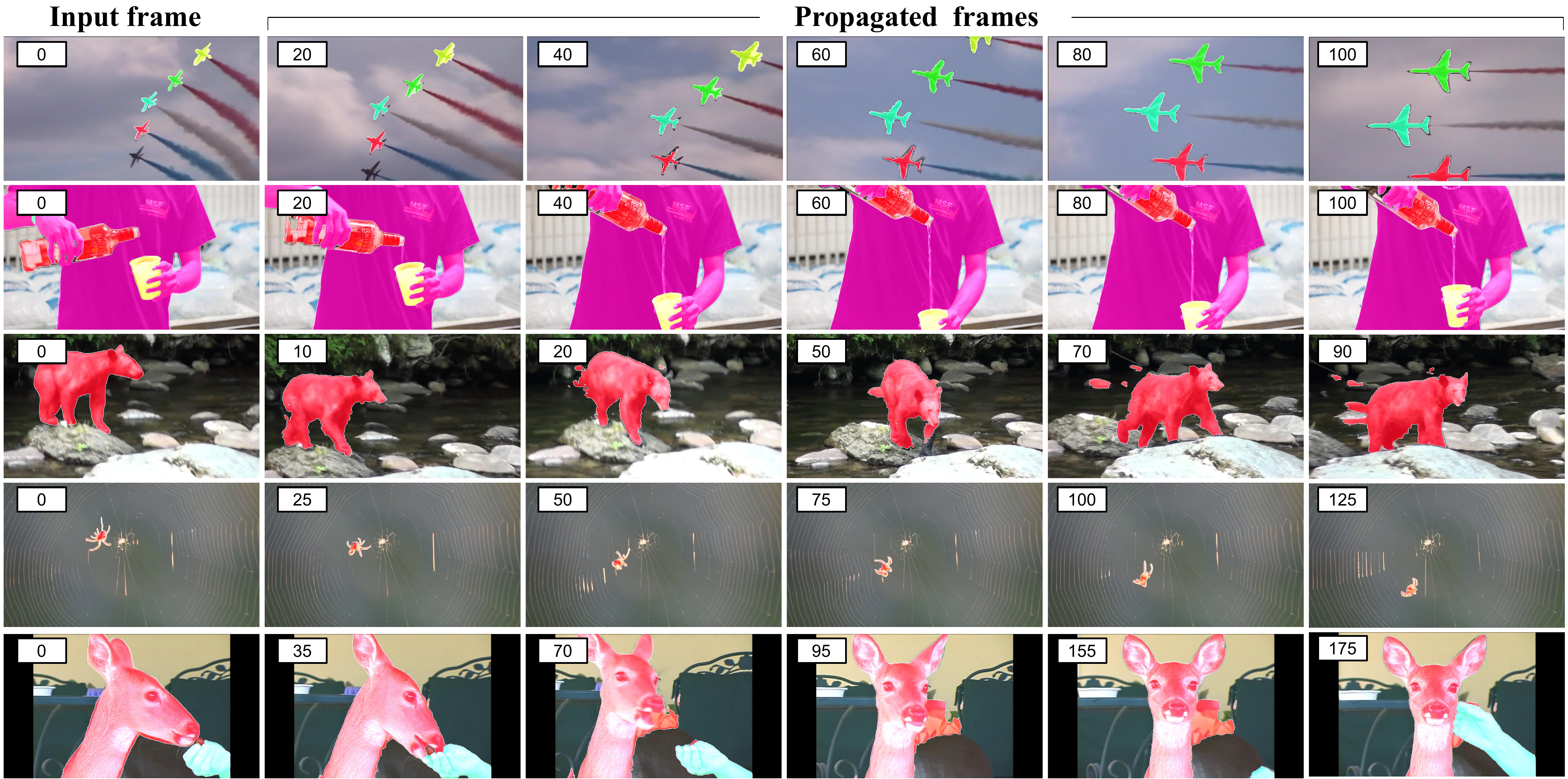}
  \captionof{figure}{\textbf{More qualitative results}  from our \emph{self-supervised dense tracking model} on the YouTube-VOS dataset. 
The number on the top left refers to the frame number in the video. Row 1: Tracking multiple similar objects with scale change. Row 2: Occlusions and out-of-scene objects (hand, bottle, and cup). Row 3: Large camera shake. Row 4: Small object with fine details. Row 5: Inferring unseen pose of the deer; out-of-scene object (hand).}
\label{fig:qualitative}
  \vspace{2em}
\end{figure*}

As shown in Figure \ref{fig:qualitative}, we provide more qualitative results exhibiting some of difficulties in the tracking task. 
These difficulties include tracking multiple similar objects (multi-instance tracking often fails by conflating similar objects), 
large camera shake (objects may have motion blur), inferring unseen object pose of objects, and so on. 
As shown in the figure, MAST handles these difficulties well.

\end{appendices}

%% file: tables/youtube-vos-19.tex

\begin{table}[!htb]
\centering
\footnotesize\addtolength{\tabcolsep}{-2pt}
\begin{tabular}{c @{\extracolsep{\fill}} ccccccc}
\toprule

\multirow{2}{*}{Method} & \multirow{2}{*}{Sup.}  & \multirow{2}{*}{Overall $\uparrow$} & \multicolumn{2}{c}{Seen} & \multicolumn{2}{c}{Unseen} & 
\multirow{2}{*}{Gen. Gap $\downarrow$}\\ 
\cmidrule{4-5} \cmidrule{6-7} 
 &  & & $\mathcal{J}$ $\uparrow$ & $\mathcal{F}$ $\uparrow$ & $\mathcal{J}$ $\uparrow$ & $\mathcal{F}$ $\uparrow$  \\ 

 \midrule

Vid. Color.\cite{Vondrick18}$^\dagger$  &  \xmark & 39.0  & 43.3  & 38.2  & 36.6 & 37.5 & 3.7 \\
CorrFlow\cite{Lai19}  &    \xmark & 47.0   & 51.2   & 46.6 & 44.5  & 45.9 &  3.7 \\
\textbf{MAST~(Ours)}   &  \xmark & \textbf{64.9}  & \textbf{64.3}  & \textbf{65.3}  & \textbf{61.5} & \textbf{68.4} &  \textbf{0.15}\\

\bottomrule
\end{tabular}
\vspace{-8pt}
\caption{Video segmentation results on Youtube-VOS 2019 dataset. Higher values are better. $^\dagger$ indicates results based on our reimplementation.}
\label{table:youtube_19}
\vspace{-2em}
\end{table}

%% file: egpaper_for_review.bbl
\begin{thebibliography}{10}\itemsep=-1pt

\bibitem{Agrawal15}
Pulkit Agrawal, Joao Carreira, and Jitendra Malik.
\newblock Learning to see by moving.
\newblock In {\em Proc. ICCV}, 2015.

\bibitem{Bahdanau2015}
Dzmitry Bahdanau, Kyunghyun Cho, and Yoshua Bengio.
\newblock Neural machine translation by jointly learning to align and
  translate.
\newblock {\em Proc. ICLR}, 2015.

\bibitem{Bao18}
Linchao Bao, Baoyuan Wu, and Wei Liu.
\newblock Cnn in mrf: Video object segmentation via inference in a cnn-based
  higher-order spatio-temporal mrf.
\newblock In {\em Proc. CVPR}, 2018.

\bibitem{Jia16}
Bert~De Brabandere, Xu Jia, Tinne Tuytelaars, and Luc~Van Gool.
\newblock Dynamic filter networks.
\newblock In {\em NIPS}, 2016.

\bibitem{Brazelton66}
T.~Berry Brazelton, Mary~Louise Scholl, and John~S. Robey.
\newblock Visual responses in the newborn.
\newblock {\em Pediatrics}, 1966.

\bibitem{Caelles17}
Sergi Caelles, Kevis-Kokitsi Maninis, Jordi Pont-Tuset, Laura Leal-Taix{\'e},
  Daniel Cremers, and Luc~Van Gool.
\newblock One-shot video object segmentation.
\newblock In {\em Proc. CVPR}, 2017.

\bibitem{Wu19}
Wu Chao-Yuan, Feichtenhofer Christoph, Fan Haoqi, He Kaiming,
  Kr\"{a}henb\"{u}hl Philipp, and Girshick Ross.
\newblock {Long-Term Feature Banks for Detailed Video Understanding}.
\newblock In {\em Proc. CVPR}, 2019.

\bibitem{Cheng18}
Jingchun Cheng, Yi-Hsuan Tsai, Wei-Chih Hung, Shengjin Wang, and Ming-Hsuan
  Yang.
\newblock Fast and accurate online video object segmentation via tracking
  parts.
\newblock In {\em Proc. CVPR}, 2018.

\bibitem{chung2014empirical}
Junyoung Chung, Caglar Gulcehre, KyungHyun Cho, and Yoshua Bengio.
\newblock Empirical evaluation of gated recurrent neural networks on sequence
  modeling.
\newblock {\em arXiv preprint arXiv:1412.3555}, 2014.

\bibitem{Deng09}
Jia Deng, Wei Dong, Richard Socher, Li-Jia Li, Kai Li, and Fei-Fei Li.
\newblock Imagenet: A large-scale hierarchical image database.
\newblock In {\em Proc. CVPR}, 2009.

\bibitem{Denton17}
Emily Denton and Vighnesh Birodkar.
\newblock Unsupervised learning of disentangled representations from video.
\newblock In {\em NIPS}, 2017.

\bibitem{Devlin18}
J. Devlin, M.W. Chang, K. Lee, and K. Toutanova.
\newblock Bert: Pre-training of deep bidirectional transformers for language
  understanding.
\newblock 2018.

\bibitem{Everingham09a}
Mark Everingham, Luc~Van Gool, Christopher K.~I. Williams, John Winn, and
  Andrew Zisserman.
\newblock The {PASCAL} {V}isual {O}bject {C}lasses {C}hallenge 2009 {(VOC2009)}
  {R}esults.
\newblock
  http://www.pascal-network.org/challenges/VOC/voc2009/workshop/index.html,
  2009.

\bibitem{Fernando17}
Basura Fernando, Hakan Bilen, Efstratios Gavves, and Stephen Gould.
\newblock Self-supervised video representation learning with odd-one-out
  networks.
\newblock In {\em Proc. CVPR}, 2017.

\bibitem{Gan18}
Chuang Gan, Boqing Gong, Kun Liu, Hao Su, and Leonidas~J. Guibas.
\newblock Geometry guided convolutional neural networks for self-supervised
  video representation learning.
\newblock In {\em Proc. CVPR}, 2018.

\bibitem{Graves2014}
Alex Graves, Greg Wayne, and Ivo Danihelka.
\newblock Neural turing machines.
\newblock {\em arXiv preprint arXiv:1410.5401}, 2014.

\bibitem{Han19}
Tengda Han, Weidi Xie, and Andrew Zisserman.
\newblock Video representation learning by dense predictive coding.
\newblock In {\em 1st International Workshop on Large-scale Holistic Video
  Understanding, ICCV}, 2019.

\bibitem{He16}
Kaiming He, Xiangyu Zhang, Shaoqing Ren, and Jian Sun.
\newblock Deep residual learning for image recognition.
\newblock In {\em Proc. CVPR}, 2016.

\bibitem{Hochreiter97}
Sepp Hochreiter and J{\"u}rgen Schmidhuber.
\newblock Long short-term memory.
\newblock {\em Neural computation}, 9(8):1735--1780, 1997.

\bibitem{Hu17}
Yuan-Ting Hu, Jia-Bin Huang, and Alexander~G. Schwing.
\newblock Maskrnn: Instance level video object segmentation.
\newblock In {\em NIPS}, 2017.

\bibitem{Hu18}
Yuan-Ting Hu, Jia-Bin Huang, and Alexander~G. Schwing.
\newblock Videomatch: Matching based video object segmentation.
\newblock In {\em Proc. ECCV}, 2018.

\bibitem{Isola15}
Phillip Isola, Daniel Zoran, Dilip Krishnan, and Edward~H. Adelson.
\newblock Learning visual groups from co-occurrences in space and time.
\newblock In {\em Proc. ICLR}, 2015.

\bibitem{Jaderberg2015}
Max Jaderberg, Karen Simonyan, Andrew Zisserman, et~al.
\newblock Spatial transformer networks.
\newblock In {\em NIPS}, 2015.

\bibitem{Jakab18}
Tomas Jakab, Ankush Gupta, Hakan Bilen, and Andrea Vedaldi.
\newblock Conditional image generation for learning the structure of visual
  objects.
\newblock In {\em NIPS}, 2018.

\bibitem{Jayaraman15}
Dinesh Jayaraman and Kristen Grauman.
\newblock Learning image representations tied to ego-motion.
\newblock In {\em Proc. ICCV}, 2015.

\bibitem{Jayaraman16}
Dinesh Jayaraman and Kristen Grauman.
\newblock Slow and steady feature analysis: higher order temporal coherence in
  video.
\newblock In {\em Proc. CVPR}, 2016.

\bibitem{Jing18}
Longlong Jing, Xiaodong Yang, Jingen Liu, and Yingli Tian.
\newblock Self-supervised spatiotemporal feature learning by video geometric
  transformations.
\newblock {\em arXiv preprint arXiv:1811.11387}, 2018.

\bibitem{Johnander19}
Joakim Johnander, Martin Danelljan, Emil Brissman, Fahad~Shahbaz Khan, and
  Michael Felsberg.
\newblock A generative appearance model for end-to-end video object
  segmentation.
\newblock In {\em Proc. CVPR}, 2019.

\bibitem{Khoreva17}
Anna Khoreva, Rodrigo Benenson, Eddy Ilg, Thomas Brox, and Bernt Schiele.
\newblock Lucid data dreaming for multiple object tracking.
\newblock In {\em arXiv preprint arXiv: 1703.09554}, 2017.

\bibitem{Khoreva16}
Anna Khoreva, Federico Perazzi, Rodrigo Benenson, Bernt Schiele, and Alexander
  Sorkine-Hornung.
\newblock Learning video object segmentation from static images.
\newblock In {\em arXiv preprint arXiv:1612.02646}, 2016.

\bibitem{Khoreva18}
A. Khoreva, A. Rohrbach, and B. Schiele.
\newblock Video object segmentation with language referring expressions.
\newblock In {\em Proc. ACCV}, 2018.

\bibitem{Kim18}
Dahun Kim, Donghyeon Cho, and In~So Kweon.
\newblock Self-supervised video representation learning with space-time cubic
  puzzles.
\newblock In {\em AAAI}, 2018.

\bibitem{Song19}
Shu Kong and Charless Fowlkes.
\newblock Multigrid predictive filter flow for unsupervised learning on videos.
\newblock {\em arXiv 1904.01693, 2019}, 2019.

\bibitem{Kremenitzer79}
Janet~P. Kremenitzer, Herbert~G. Vaughan, Diane Kurtzberg, and Kathryn Dowling.
\newblock Smooth-pursuit eye movements in the newborn infant.
\newblock {\em Child Development}, 1979.

\bibitem{VOT_TPAMI}
Matej Kristan, Jiri Matas, Ale\v{s} Leonardis, Tomas Vojir, Roman Pflugfelder,
  Gustavo Fernandez, Georg Nebehay, Fatih Porikli, and Luka \v{C}ehovin.
\newblock A novel performance evaluation methodology for single-target
  trackers.
\newblock {\em IEEE Transactions on Pattern Analysis and Machine Intelligence},
  2016.

\bibitem{Kumar2016}
Ankit Kumar, Ozan Irsoy, Peter Ondruska, Mohit Iyyer, James Bradbury, Ishaan
  Gulrajani, Victor Zhong, Romain Paulus, and Richard Socher.
\newblock Ask me anything: Dynamic memory networks for natural language
  processing.
\newblock In {\em International conference on machine learning}, 2016.

\bibitem{Lai19}
Zihang Lai and Weidi Xie.
\newblock Self-supervised learning for video correspondence flow.
\newblock In {\em Proc. BMVC.}, 2019.

\bibitem{Lee17}
Hsin-Ying Lee, Jia-Bin Huang, Maneesh Singh, and Ming-Hsuan Yang.
\newblock Unsupervised representation learning by sorting sequences.
\newblock In {\em Proc. ICCV}, 2017.

\bibitem{Li18ECCV}
Xiaoxiao Li and Chen~Change Loy.
\newblock Video object segmentation with joint re-identification and
  attention-aware mask propagation.
\newblock In {\em Proc. ECCV}, 2018.

\bibitem{Lin14}
Tsung-Yi Lin, Michael Maire, Serge Belongie, Lubomir Bourdev, Ross Girshick,
  James Hays, Pietro Perona, Deva Ramanan, C.~Lawrence Zitnick, and Piotr
  Dollar.
\newblock Microsoft coco: Common objects in context.
\newblock In {\em Proc. ECCV}, 2014.

\bibitem{Luiten18}
Jonathon Luiten, Paul Voigtlaender, and Bastian Leibe.
\newblock Premvos: Proposal-generation, refinement and merging for video object
  segmentation.
\newblock In {\em Proc. ACCV}, 2018.

\bibitem{Maninis18}
Kevis-Kokitsi Maninis, Sergi Caelles, Yuhua Chen, Jordi Pont-Tuset, Laura
  Leal-Taix\'e, Daniel Cremers, and Luc {Van Gool}.
\newblock Video object segmentation without temporal information.
\newblock {\em IEEE Transactions on Pattern Analysis and Machine Intelligence
  (TPAMI)}, 2018.

\bibitem{Misra16}
Ishan Misra, C.~Lawrence Zitnick, and Martial Hebert.
\newblock Shuffle and learn: Unsupervised learning using temporal order
  verification.
\newblock In {\em Proc. ECCV}, 2016.

\bibitem{Oh18}
Seoung~Wug Oh, Joon-Young Lee, Kalyan Sunkavalli, and Seon~Joo Kim.
\newblock Fast video object segmentation by reference-guided mask propagation.
\newblock In {\em Proc. CVPR}, 2018.

\bibitem{Seoung19}
Seoung~Wug Oh, Joon-Young Lee, Ning Xu, and Seon~Joo Kim.
\newblock Video object segmentation using space-time memory networks.
\newblock In {\em Proc. ICCV}, 2019.

\bibitem{Oord16}
Aaron van~den Oord, Nal Kalchbrenner, and Koray Kavukcuoglu.
\newblock Pixel recurrent neural networks.
\newblock In {\em Proc. ICML}, 2016.

\bibitem{Perazzi16}
F. Perazzi, J. Pont-Tuset, B. McWilliams, L. {Van Gool}, M. Gross, and A.
  Sorkine-Hornung.
\newblock A benchmark dataset and evaluation methodology for video object
  segmentation.
\newblock In {\em Proc. CVPR}, 2016.

\bibitem{Pont-Tuset17}
Jordi Pont-Tuset, Federico Perazzi, Sergi Caelles, Pablo Arbeláez, Alex
  Sorkine-Hornung, and Luc~Van Gool.
\newblock The 2017 davis challenge on video object segmentation.
\newblock {\em arXiv:1704.00675}, 2017.

\bibitem{Reinhard01}
Erik Reinhard, Michael Adhikhmin, Bruce Gooch, and Peter Shirley.
\newblock Color transfer between images.
\newblock {\em IEEE Computer graphics and applications}, 21(5):34--41, 2001.

\bibitem{See2017}
Abigail See, Peter~J Liu, and Christopher~D Manning.
\newblock Get to the point: Summarization with pointer-generator networks.
\newblock {\em arXiv preprint arXiv:1704.04368}, 2017.

\bibitem{Sukhbaatar2015}
Sainbayar Sukhbaatar, Jason Weston, Rob Fergus, et~al.
\newblock End-to-end memory networks.
\newblock In {\em Advances in neural information processing systems}, pages
  2440--2448, 2015.

\bibitem{Valmadre18}
Jack Valmadre, Luca Bertinetto, Joao~F. Henriques, Ran Tao, Andrea Vedaldi,
  Arnold Smeulders, Philip Torr, and Efstratios Gavves.
\newblock Long-term tracking in the wild: A benchmark.
\newblock In {\em Proc. ECCV}, 2018.

\bibitem{Vaswani17}
A. Vaswani, N. Shazeer, N. Parmar, J. Uszkoreit, L. Jones, A.~N. Gomez, \L.
  Kaiser, and I. Polosukhin.
\newblock Attention is all you need.
\newblock In {\em NIPS}, 2017.

\bibitem{Ventura19}
Carles Ventura, Miriam Bellver, Andreu Girbau, Amaia Salvador, Ferran Marques,
  and Xavier Giro-i Nieto.
\newblock Rvos: End-to-end recurrent network for video object segmentation.
\newblock In {\em Proc. CVPR}, 2019.

\bibitem{Ventura_2019_CVPR}
Carles Ventura, Miriam Bellver, Andreu Girbau, Amaia Salvador, Ferran Marques,
  and Xavier Giro-i Nieto.
\newblock Rvos: End-to-end recurrent network for video object segmentation.
\newblock In {\em Proc. CVPR}, June 2019.

\bibitem{Voigtlaender19}
Paul Voigtlaender, Yuning Chai, Florian Schroff, Hartwig Adam, Bastian Leibe,
  and Liang-Chieh Chen.
\newblock Feelvos: Fast end-to-end embedding learning for video object
  segmentation.
\newblock In {\em Proc. CVPR}, 2019.

\bibitem{Voigtlaender17}
Paul Voigtlaender and Bastian Leibe.
\newblock Online adaptation of convolutional neural networks for video object
  segmentation.
\newblock In {\em Proc. BMVC.}, 2017.

\bibitem{Hofsten82}
C. von Hofsten.
\newblock Eye–hand coordination in the newborn.
\newblock {\em Developmental Psychology}, 1982.

\bibitem{Vondrick18}
Carl Vondrick, Abhinav Shrivastava, Alireza Fathi, Sergio Guadarrama, and Kevin
  Murphy.
\newblock Tracking emerges by colorizing videos.
\newblock In {\em Proc. ECCV}, 2018.

\bibitem{Wang19_Tracking}
Ning Wang, Yibing Song, Chao Ma, Wengang Zhou, Wei Liu, and Houqiang Li.
\newblock Unsupervised deep tracking.
\newblock In {\em Proc. CVPR}, 2019.

\bibitem{Wang19a}
Qiang Wang, Li Zhang, Luca Bertinetto, Weiming Hu, and Philip~H.S. Torr.
\newblock Fast online object tracking and segmentation: A unifying approach.
\newblock In {\em Proc. CVPR}, 2019.

\bibitem{Wang2018}
Xiaolong Wang, Ross Girshick, Abhinav Gupta, and Kaiming He.
\newblock Non-local neural networks.
\newblock In {\em Proc. CVPR}, 2018.

\bibitem{Wang15}
Xiaolong Wang and Abhinav Gupta.
\newblock Unsupervised learning of visual representations using videos.
\newblock In {\em Proc. ICCV}, 2015.

\bibitem{Wang19}
Xiaolong Wang, Allan Jabri, and Alexei~A. Efros.
\newblock Learning correspondence from the cycle-consistency of time.
\newblock In {\em Proc. CVPR}, 2019.

\bibitem{Wang19RA}
Ziqin Wang, Jun Xu, Li Liu, Fan Zhu, and Ling Shao.
\newblock Ranet: Ranking attention network for fast video object segmentation.
\newblock In {\em Proc. ICCV}, 2019.

\bibitem{Wei18}
Donglai Wei, Joseph Lim, Andrew Zisserman, and William~T. Freeman.
\newblock Learning and using the arrow of time.
\newblock In {\em Proc. CVPR}, 2018.

\bibitem{Wiles18a}
Olivia Wiles, A.~Sophia Koepke, and Andrew Zisserman.
\newblock Self-supervised learning of a facial attribute embedding from video.
\newblock In {\em Proc. BMVC.}, 2018.

\bibitem{Wiles18}
Olivia Wiles, A.~Sophia Koepke, and Andrew Zisserman.
\newblock X2face: A network for controlling face generation using images,
  audio, and pose codes.
\newblock In {\em Proc. ECCV}, 2018.

\bibitem{Zhu17}
Zhu Xizhou, Wang Yujie, Dai Jifeng, Yuan Lu, and Wei Yichen.
\newblock Flow-guided feature aggregation for video object detection.
\newblock In {\em Proc. ICCV}, 2017.

\bibitem{Xu2015}
Kelvin Xu, Jimmy Ba, Ryan Kiros, Kyunghyun Cho, Aaron Courville, Ruslan
  Salakhudinov, Rich Zemel, and Yoshua Bengio.
\newblock Show, attend and tell: Neural image caption generation with visual
  attention.
\newblock In {\em Proc. ICML}, 2015.

\bibitem{Xu18}
Ning Xu, Linjie Yang, Yuchen Fan, Dingcheng Yue, Yuchen Liang, Jianchao Yang,
  and Thomas Huang.
\newblock Youtube-vos: A large-scale video object segmentation benchmark.
\newblock {\em arXiv:1809.03327}, 2018.

\bibitem{Li19}
Li Xueting, Liu Sifei, De~Mello Shalini, Wang Xiaolong, Kautz Jan, and Yang
  Ming-Hsuan.
\newblock Joint-task self-supervised learning for temporal correspondence.
\newblock In {\em NeurIPS}, 2019.

\bibitem{Yang18}
Linjie Yang, Yanran Wang, Xuehan Xiong, Jianchao Yang, and Aggelos Katsaggelos.
\newblock Efficient video object segmentation via network modulation.
\newblock In {\em Proc. CVPR}, 2018.

\bibitem{Yao2015}
Li Yao, Atousa Torabi, Kyunghyun Cho, Nicolas Ballas, Christopher Pal, Hugo
  Larochelle, and Aaron Courville.
\newblock Describing videos by exploiting temporal structure.
\newblock In {\em Proc. ICCV}, 2015.

\bibitem{Zhang16_Color}
Richard Zhang, Phillip Isola, and Alexei Efros.
\newblock Colorful image colorization.
\newblock In {\em Proc. ECCV}, 2016.

\end{thebibliography}


\begin{thebibliography}{1}

\bibitem{Brock19}
A.~Brock, J.~Donahue, and K.~Simonyan.
\newblock Large scale gan training for high fidelity natural image synthesis.
\newblock In {\em ICML}, 2019.

\bibitem{Isola15}
P.~Isola, D.~Zoran, D.~Krishnan, and E.~H. Adelson.
\newblock Learning visual groups from co-occurrences in space and time.
\newblock In {\em Proc. ICLR}, 2015.

\bibitem{Kolesnikov19}
A.~Kolesnikov, X.~Zhai, and L.~Beyer.
\newblock Revisiting self-supervised visual representation learning.
\newblock In {\em CVPR}, 2019.

\end{thebibliography}
